\documentclass[journal]{IEEEtran}

\ifCLASSOPTIONcompsoc
  \usepackage[nocompress]{cite}
\else
  \usepackage{cite}
\fi

\usepackage{graphicx}
\usepackage{makecell}
 \usepackage{diagbox}
 \usepackage{adjustbox}
\usepackage[utf8]{inputenc}

\usepackage[english]{babel}
\usepackage[utf8]{inputenc} 
\usepackage[bookmarks=false,hyperfootnotes=false]{hyperref}
\hypersetup{
			colorlinks=true,
			linkcolor=blue,
			anchorcolor=black,
			citecolor=green,
			urlcolor=blue
			}
\usepackage{tipa}
 \expandafter\let\csname equation*\endcsname\relax
  \expandafter\let\csname endequation*\endcsname\relax
\usepackage{amsmath,amsfonts,amssymb}

\usepackage{amsthm}
\usepackage{mathtools}

\newcommand{\RR}{\mathbb{R}}

\newcommand{\Dcal}{\mathcal{D}}
\newcommand{\Ecal}{\mathcal{E}}

\newcommand{\Ocal}{\mathcal{O}}

\newcommand{\Rcal}{\mathcal{R}}

\newcommand{\Ibold}{\mathbf{I}}

\newcommand{\Mbold}{\mathbf{M}}

\newcommand{\Wbold}{\mathbf{W}}

\newcommand{\dbold}{{\bm{d}}}

\newcommand{\nbold}{{\bm{n}}}

\newcommand{\xbold}{{\bm{x}}}
\newcommand{\ybold}{{\bm{y}}}

\newcommand{\zbold}{{\bm{z}}}

\usepackage{bm} 

\usepackage[normalem]{ulem}

\DeclareMathOperator*{\argmin}{argmin}

\AtBeginDocument{
  
  \let\argmin\relax
  \let\diag\relax

  \let\div\relax

  \let\lap\relax

  \let\prox\relax

  \DeclareMathOperator*{\argmin}{arg\,min}
  \DeclareMathOperator{\diag}{diag}

  \DeclareMathOperator{\div}{div}

  \DeclareMathOperator{\lap}{\Delta}

  \DeclareMathOperator{\prox}{\mathbf{prox}}

}

\usepackage[font=scriptsize]{subcaption}

\usepackage{bm}
\newcommand{\grad}{{\bm{\nabla}}}
\usepackage{soul}

\newcommand{\blank}{\,{\cdot}\,}

\usepackage{tabularx}
\usepackage{booktabs}
\usepackage{multirow}
\usepackage{algorithmicx}
\usepackage[titlenumbered,ruled,vlined]{algorithm2e}
\SetKw{Break}{break}
\usepackage{algpseudocode,float}
\algnewcommand\Input{\textbf{Input: }}
\algnewcommand\Package{\textbf{Package: }}
\algnewcommand\Parameters{\textbf{Parameters: }}
\algnewcommand\Output{\textbf{Output: }}

\usepackage{xparse}
   \DeclareDocumentCommand\TV{mmo}
 {
  \IfValueTF{#3}
   {\mathrm{{TV}}_{{#2}}^{#1}({#3})}
   {\mathrm{{TV}}_{{#2}}^{#1}}%
 }

   \DeclareDocumentCommand\TDVM{mmo}
 {
  \IfValueTF{#3}
   {\mathrm{{TDV}}_{{#2}}^{#1}({#3})}
   {\mathrm{{TDV}}_{{#2}}^{#1}}%
 }

\DeclareDocumentCommand\LL{mo}
 {
  \IfValueTF{#2}
   {\mathrm{{L}}^{#1}(#2)}
   {\mathrm{{L}}^{#1}}%
 }
\newcommand{\T}{\mathrm{T}} 

\newcommand{\diff}{\mathop{}\mathrm{d}}

\providecommand*{\pderiv}[3][]{\dfrac{\partial^{#1}#2}
{\partial #3^{#1}}}

\newcommand{\abs}[1]{\left\lvert #1 \right\rvert}
\newcommand{\norm}[1]{\left\lVert #1 \right\rVert}

\newcommand{\cbs}[1]{\textcolor{green}{Carola: #1}}

\newcommand{\comment}[1]{}
\allowdisplaybreaks[1]
\algnewcommand{\IfThenElse}[3]{
  \algorithmicif\ #1\ \algorithmicthen\ #2\ \algorithmicelse\ #3}
 \algnewcommand{\IfThen}[2]{
  \algorithmicif\ #1\ \algorithmicthen\ #2}

\providecommand*{\eu}{\ensuremath{\mathrm{e}}} 
\newcommand{\rev}[1]{#1}
\newcommand{\revbis}[1]{#1}

\begin{document}

\title{Variational Osmosis for Non-linear Image Fusion}

\author{Simone~Parisotto,
        Luca~Calatroni,
        Aurelie~Bugeau,
        Nicolas~Papadakis
        and Carola-Bibiane~Sch\"{o}nlieb.
\thanks{S.\ Parisotto and C.-B.\ Sch\"{o}nlieb are with DAMTP, University of Cambridge, Cambridge CB3 0WA, UK. Correspondence e-mail: \url{sp751@cam.ac.uk}
}
\thanks{L.\ Calatroni is with CNRS, Université Côte d’Azur,  INRIA, I3S, UMR 7271, Sophia-Antipolis, France }
\thanks{A.\ Bugeau is with Univ.\ Bordeaux, Bordeaux INP, CNRS, LaBRI, UMR 5800, F-33400 Talence, France}
\thanks{N.\ Papadakis is with CNRS, Univ.\ Bordeaux, IMB, UMR 5251, F-33400 Talence, France}
}

\IEEEtitleabstractindextext{%
\begin{abstract}
We propose a new variational model for non-linear image fusion.  \rev{Our approach is based on the use of an osmosis energy term related to the one studied}  in Vogel et al.~ \cite{Vogel2013} and Weickert et al.~ \cite{Weickert2013}. The \rev{minimization of the proposed non-convex energy realizes} visually plausible image data fusion, \rev{invariant to multiplicative brightness changes}. On the practical side, it requires minimal supervision and parameter tuning and can encode prior information on the structure of the images to be fused. 
For the numerical solution of the proposed model, we develop a primal-dual algorithm and we apply the resulting minimization scheme to solve multi-modal face fusion, color transfer and cultural heritage conservation problems. Visual \rev{and quantitative} comparisons to state-of-the-art approaches prove the \rev{out-performance} and the flexibility of our method. 
\end{abstract}

}

\maketitle              

\IEEEdisplaynontitleabstractindextext

\section{Introduction}
Image fusion is broadly referred to as the problem of gluing information from two (or more) images so as to create a composite image combining the structures of the images used in a pleasant way.  A recent survey \cite{Li2017} has shown the increasing interest for image fusion for a wide range applications such as image editing~\cite{Perez2003}, image enhancement~\cite{Ancuti2012enhancing} focus-stacking imaging \cite{Li2008multifocus}, HDR \rev{(\emph{High-Dynamic-Range})} imaging~\cite{Mertens2007exposure}, dehazing~\cite{Ancuti2016night}, facial texture transfers~\cite{Thies2015real}, etc. Moreover, image fusion can be used as a technique to unveil hidden information visible only by means of multi-modal image modalities, see, e.g.\ \cite{Parisotto2018,Calatroni2018} for applications to cultural heritage. In the context of medical imaging, image fusion enables to combine different modalities for better diagnosis~\cite{James2014medical,Saad2013}.  This is also useful in remote sensing~\cite{Ma2019infrared} where visible light from standard camera sensors can be combined with spectral imaging bands such as infrared. \revbis{In the same context, image fusion can be further used as an inpainting-type technique for photo editing or occluded object removal, e.g.\ as in the case of the removal of clouds in multi-temporal registered satellite images \cite{Gabarda2007}.}
%
\vspace{-1em}
\subsection{ \revbis{Related works}}
\rev{
Classical approaches for image fusion are typically formulated in the spatial image domain The most classical ones are based on averaging, principal component analysis and spectral embedding, see, e.g.\ \cite{Mishra2015image,Zhang2017}. Multi-scale decomposition techniques for image fusion can be formulated either in the spatial or in the spectral domain. Examples are approaches based on wavelets \cite{PAJARES20041855,Kim2011}, Laplacian pyramid decomposition \cite{Burt83} or sparse representations \cite{Zhao2018}. We refer the reader to \cite{Amolins2007} for a general review. 
}

\rev{
Over the last twenty years, mathematical models for image fusion based on variational techniques have also been proposed. Some of them are based on the optimization of perceptually-inspired measures of local contrast and illumination (see, e.g., \cite{HafnerPHD,Tian2018}), while others look directly at structural image information (such as edges and curvature) of both images, trying to merge them by means of sparsity-promoting regularizers \cite{Li2016}. In \cite{Zhang2020}, the authors consider a space-variant extension of this approach in order to improve fusion results for remote sensing applications.
}

\rev{
Alternatively, mathematical approaches based on the use of first and second order Partial Differential Equations (PDEs) can be considered. Among them, a very popular one is the well-known Poisson image editing \cite{Perez2003}. There, the fused image is computed by minimizing the difference between its gradient and the one of the plain gluing of the two images to-be-fused in the subregion of the image domain where the fusion is desired. 
Such problem can be equivalently formulated in terms of a linear elliptic PDE, endowed with appropriate boundary conditions. In terms of evolution models, a linear isotropic drift-diffusion PDE describing the physical phenomenon of osmosis \cite{Hagenburg2012} has been applied in \cite{Weickert2013} to solve the image fusion problem. Compared to standard Poisson editing, the main advantage of the linear osmosis model is its invariance to multiplicative brightness changes, which makes the fusion much more effective in the presence of images with different contrast. Furthermore, the osmosis model can be easily applied to other problems such as image compression and shadow removal, and it is amenable for efficient numerical schemes guaranteeing mass and positiveness preservation of the solution, see \cite{Vogel2013,Calatroni2017}. In order to overcome the smoothing artefacts of the linear model, an anisotropic osmosis variant driving image diffusion along coherent directions has been considered in \cite{Parisotto_2019}.
There, the reference drift estimation and the image reconstruction step are performed in a disjoint, yet still linear, manner.}


\rev{
The use of non-linear models for image fusion has recently been employed in \cite{BenMolNosBurCreGilSch2017,Hait2019}, where the non-linear spectral decomposition of the Total Variation (TV) regularization functional is used to improve texture fusion in a multi-scale scheme. The performance of such approach is, however, somehow limited by the manual pre-processing (e.g.\ the image alignment step) and careful parameter choice (such as the one of the specific eigenvectors extracted from both images capturing fine details with sufficient precision) required.
}
\smallskip
\subsubsection*{Contribution}
\begin{figure}
    \centering
    \includegraphics[width=0.495\textwidth]{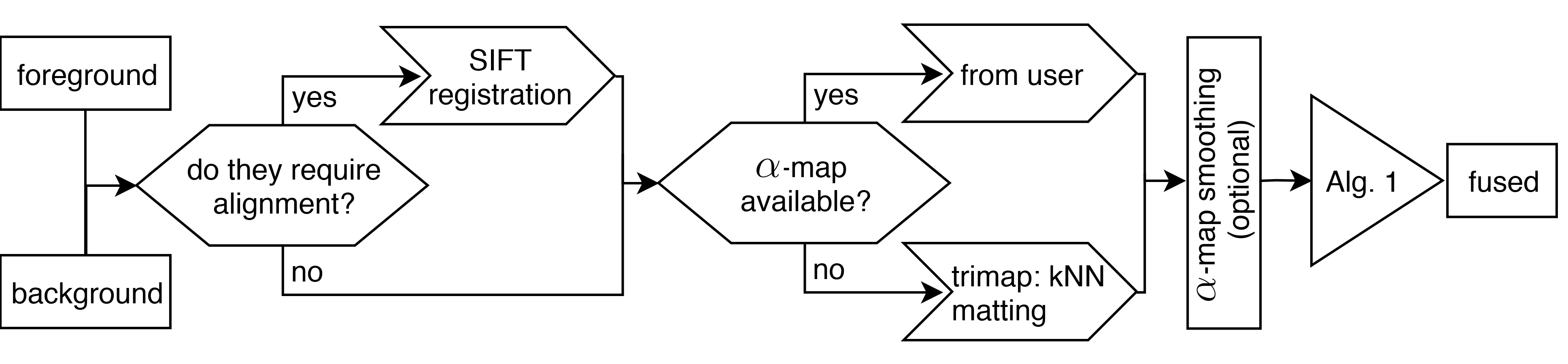}
    \caption{\rev{Flowchart of the proposed model for image fusion.}}
    \label{fig: flowchart}
\end{figure}

\begin{figure}[ht!]
    \centering
    \begin{subfigure}[t]{0.2355\textwidth}\centering
    \captionsetup{justification=centering}
    \includegraphics[width=0.99\textwidth,trim=13cm 6.75cm 2.8cm 31cm,clip=true]{{./results_vr/2_foreground_low}.jpg}
    \caption{Visible (foreground $f$).}
    \end{subfigure}
    \hfill
    \begin{subfigure}[t]{0.2355\textwidth}\centering
    \captionsetup{justification=centering}
    \includegraphics[width=0.99\textwidth,trim=13cm 6.75cm 2.8cm 31cm,clip=true]{{./results_vr/2_background_low}.jpg}
    \caption{Infrared (background $b$).}
    \end{subfigure}
    \\
    \begin{subfigure}[t]{0.2355\textwidth}\centering
    \captionsetup{justification=centering}
    \includegraphics[width=0.99\textwidth,trim=13cm 6.75cm 2.8cm 31cm,clip=true]{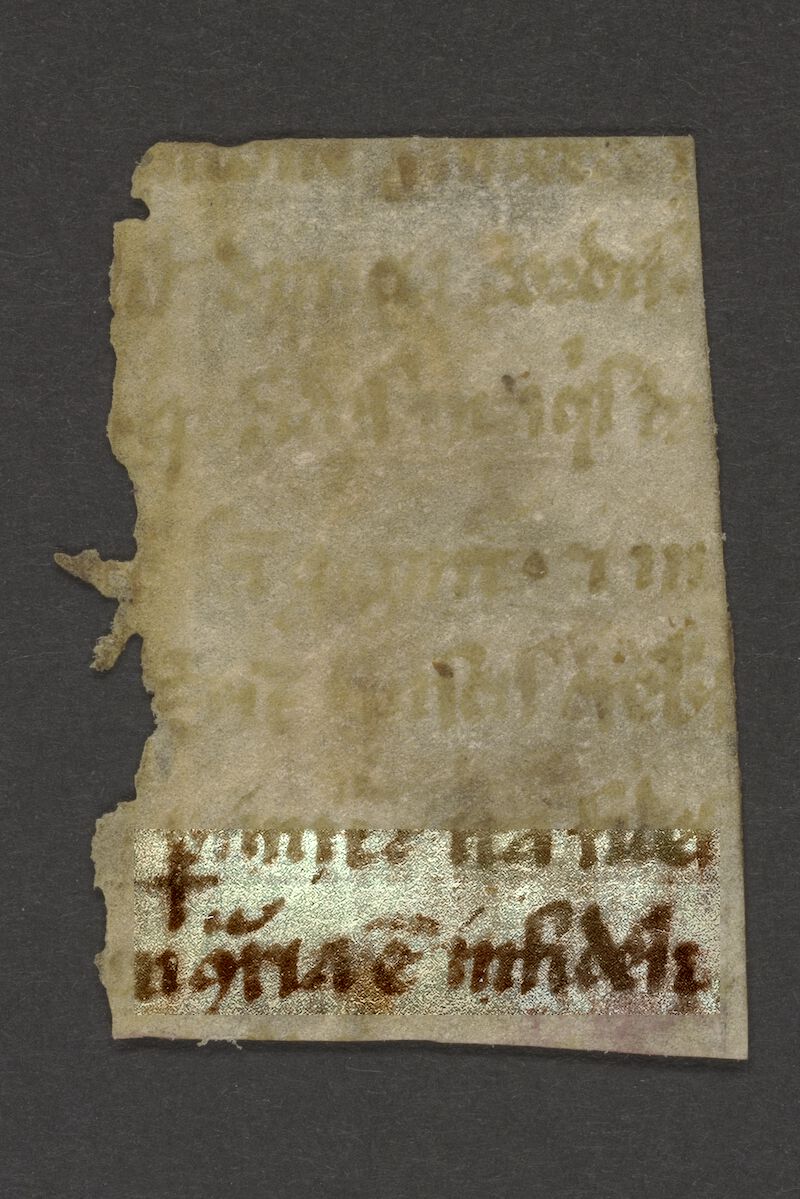}
    \caption{Poisson seamless cloning \cite{Perez2003}.}
    \end{subfigure}
    \hfill
    \begin{subfigure}[t]{0.2355\textwidth}\centering
    \captionsetup{justification=centering}
    \includegraphics[width=0.99\textwidth,trim=2.6cm 3.25cm 0.85cm 2.25cm,clip=true]{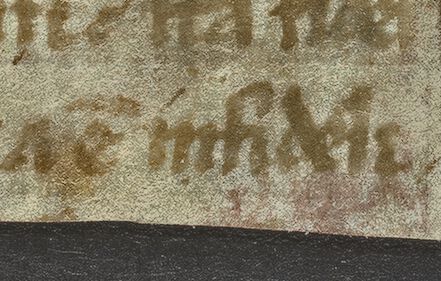}
    \caption{\rev{Non-linear Spectral TV \cite{BenMolNosBurCreGilSch2017}.}}
    \end{subfigure}
    \\
    \begin{subfigure}[t]{0.2355\textwidth}\centering
    \captionsetup{justification=centering}
    \includegraphics[width=0.99\textwidth,trim=2.6cm 3.25cm 0.85cm 2.25cm,clip=true]{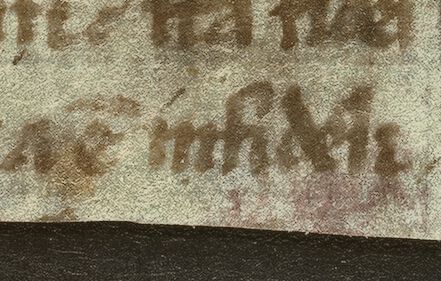}
    \caption{\rev{Classic Osmosis \cite{Weickert2013}.}}
    \end{subfigure}
    \hfill
    \begin{subfigure}[t]{0.2355\textwidth}\centering
    \captionsetup{justification=centering}
    \includegraphics[width=0.99\textwidth,trim=10.3cm 5.25cm 2.25cm 24.75cm,clip=true]{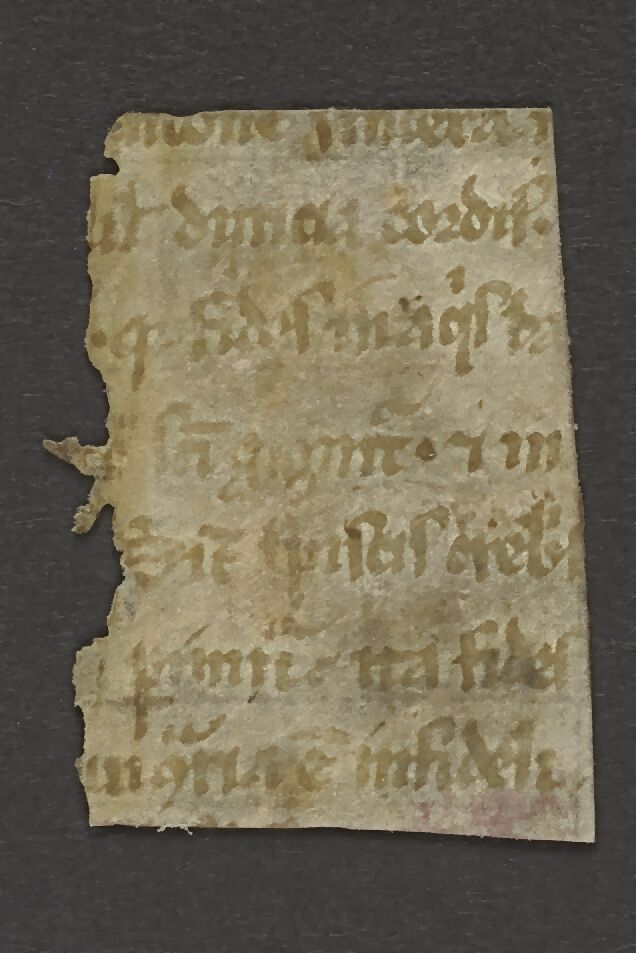}
    \caption{Proposed variational osmosis.}
    \end{subfigure}
    \caption{Multi-modal fusion of Cultural Heritage images. The structures of the infrared image (b) are transferred to the visible one (a).  Structural regularisation and chromatic preservation is enhanced using our non-linear model \rev{(f) 
    compared to state of the art methods (c) \cite{Perez2003}, (d) \cite{BenMolNosBurCreGilSch2017} and (e) \cite{Weickert2013}.}}
    \label{fig: intro example}
\end{figure}

We propose a non-linear image fusion \rev{method} formulated in terms of a non-convex variational model \rev{(flowchart in Figure \ref{fig: flowchart})}. The proposed method improves upon the aforementioned drawbacks both of the linear osmosis model and of non-linear spectral methods, \rev{as it fuses the two images while balancing the colors or recoloring them at the same time.} Our model requires little supervision and parameter tuning since it performs the estimation of the drift image (driving the osmosis regularisation) and the corresponding fusion jointly.  It enhances chromatic consistency with the image to-be-fused, favouring texture preservation and possible data smoothing, see Figure \ref{fig: intro example}.

\rev{In order to provide a quantitative assessment of our model, we also define a novel error measure,  called Geometric Chromaticity Mean (GCM) which quantifies at the same time structure and color changes after the fusion process. A comparison with classically-used quality measures shows that this new criterion better adapts to assess the quality of algorithms for image fusion and that it further agrees with visually perceived results.}

In our model, the fused image $u$ is obtained from two given foreground and background images, respectively denoted by $f$ and $b$. Their structures are forced to be glued together by means of a dedicated image $v$ which is defined in terms of a mixing $\alpha$-map (see Section \ref{sec: variational model}). 
The solution of the model is computed as a minimizing pair $(u,v)$ of the energy functional 
\begin{equation}\label{eq: joint model final}
\min_{u,v} ~ \left\{ \Ecal(u,v):=\Ocal(u,v)+\gamma\Dcal(u)+\eta\Rcal(v) \right\},
\end{equation}
where $\Ocal$ \rev{is} an \emph{osmosis} term promoting image fusion, $\Dcal$ is a fidelity term forcing $u$ to stay close to $f$ in the foreground region
and $\Rcal$ acts as a regularizer enhancing the salient information of the structural image $v$\rev{: these latter terms} are weighted by non negative parameters $\gamma$ and $\eta$ and their choice is detailed in Section \ref{sec: variational model}.

Our model allows for more flexibility than the one proposed in \cite{BenMolNosBurCreGilSch2017} as \rev{it takes} into account possible prior information available on the structures to-be-fused (such as, for instance piece-wise constant images) encoded in $v$.
\comment{
When no prior information is available, its weight $\eta$ may be simply set to zero.

We formulate the problem in a primal-dual form and 
}
To compute efficiently the numerical solution of \eqref{eq: joint model final}, we use dedicated non-convex optimization algorithms.
Numerical results show that the proposed model outperforms state-of-the-art linear and non-linear methods for face fusion, color transfer and cultural heritage imaging applications in terms of chromaticity errors and texture preservation.
\smallskip
\subsubsection*{Organisation of the paper}
In Section \ref{sec: osmsosis review} we review some classical methods for image fusion formulated in terms of a variational problem or of PDE, highlighting their limitations mainly in terms of over-smoothing and misalignment artefacts.
In Section \ref{sec: variational model} we introduce the proposed variational osmosis model and discuss on its analytical interpretation.
In Section \ref{sec: numerical details} we describe the numerical details for the computational solution of the model which is used in Section \ref{sec: applications} in a variety of data fusion applications, such as face fusion and multi-modal fusion from data acquired in Cultural Heritage conservation.

\section{Variational and PDE models for image fusion}\label{sec: osmsosis review}

\subsubsection*{Poisson editing} A very simple  classical model for image fusion is the Poisson editing model proposed by P{\'e}rez et al. in \cite{Perez2003}. The model aims at fusing areas of two different images in a visually plausible way. In detail, given two color images \revbis{$f\in H^1(\Omega;\RR^3_+)$ and $b\in L^\infty(\Omega;\RR^3_+)$ defined on the same closed image domain $\Omega$, the model copies a (not necessarily connected) closed sub-region $A\subset\Omega\subset\RR^2$ with boundary $\partial A$ of $f$ over $b$} 
by means of a guidance field $\dbold$. 
\revbis{When $\dbold:=\grad v\in L^2(\Omega; (\RR^3)^2)$ is chosen as the gradient of a source function $v\in H^1(\Omega;\RR^3_+)$}, and by taking $v=f$ as in \cite{Perez2003}, the model can be mathematically expressed channel-wise in the following variational form:
find $u^*$ such that
\[
u^*=
\rev{\argmin_{u\in H^1(\Omega)\,:\, u_{\Omega\setminus A}=b_{\Omega\setminus A}}}~ \int_A|\grad u(\xbold) -\grad v(\xbold)|^2\diff\xbold,
\]
where the constraint 
forces the matching of the resulting image $u^*$ with $b$ outside $A$, while driving the fusion inside $A$ using the structural information coming from $f$. Note that in the original formulation of the Poisson editing problem, the constraints are imposed only on $\partial A$ as a boundary condition for the corresponding PDE. However, we extend here the constraint to draw more connections with our proposed model, as it will be clear in the following.
The problem above can be formulated over the whole domain $\Omega$ as:
\rev{
\[
u^*=\argmin_{u\in H^1(\Omega)\,:\,  u_{\Omega\setminus A}=b_{\Omega\setminus A}} \int_\Omega \chi_{A}(\xbold)|\grad u(\xbold) -\grad v(\xbold)|^2\diff\xbold,
\]
}
where $\chi_{A}(\blank)$ is the characteristic function of the subset $A$ \rev{(i.e., $\chi_{A}(\xbold)=1$ if $\xbold\in A$ and $\chi_{A}(\xbold)=0$ otherwise).}


By computing the correspondent Euler-Lagrange equation, the model can be equivalently rewritten by means of a linear elliptic Poisson PDE \revbis{in weak form}, so as to solve:
\[
\text{find }\ u^*\quad \text{ s.t. }\quad\Delta u^* = \Delta v\quad\text{on }A\text{ and } u_{\Omega\setminus A}=b_{\Omega\setminus A}.
\]
The solution $u^*$ of the problem above can be equivalently computed as the steady state of the parabolic Poisson editing PDE, which reads
\[
\partial_t u=\div(\chi_A(\grad u-\grad v)),\quad\text{on }\Omega \times (0, T],
\]
which corresponds to the steepest descent formulation of the variational problem above on the whole $\Omega$.

This simple model has been successfully used for several seamless cloning applications, such as image insertion or feature exchange. Poisson editing falls into the category of gradient domain methods (see also \cite{Fattal2002}) and shares with them the capability of recovering $v$ inside $A$ up to an additive constant. This property constitutes in fact one of the main limitations of these models when applied to the problem of image cloning. By allowing for any color channel only additive shifts, Poisson editing is not able to fuse a low contrast image into a high contrast one, see Figure \ref{fig: face-fusion seamless}.

\smallskip
\subsubsection*{Image osmosis}
A more flexible model allowing for multiplicative image variations and applicable to many more problems (shadow removal, image decompression and many more) is the parabolic image osmosis model originally proposed in \cite{Weickert2013} and later considered \rev{in \cite{Vogel2013, Calatroni2017, Parisotto_2019}}. Given a final stopping time $T>0$, \revbis{two images $u^0\in L^\infty(\Omega;\RR^3_+), v\in H^1(\Omega;\RR^3_+)$, \rev{the \emph{drift} vector field $\dbold:=\grad\log v \in L^2(\Omega;(\RR^3)^2)$.}}
and a positive definite symmetric matrix  $\Wbold: \Omega \to \RR^2$, the convection-diffusion PDE osmosis model for \revbis{$u\in H^1((0,T]; H^1(\Omega;\RR^3_+))$} reads:
\begin{equation}
\begin{cases}
\label{eq: equation directional}
\rev{\partial_t u} = \div\left(\Wbold (\grad u - \dbold u) \right) &\text{on } \Omega\times(0,T],\\
u(\xbold,0) = u^0(\xbold) &\text{on } \Omega,\\
\langle\, \Wbold  \left( \grad u - \dbold u\right), \nbold \,\rangle =0 &\text{on } \partial\Omega \times (0,T],
\end{cases}
\end{equation}
where $\nbold$ denotes the outer normal vector
on $\partial\Omega$.

In \cite{Hagenburg2012,Weickert2013}, the authors looked both at the statistical and at the theoretical foundation of the model in its isotropic form, i.e.\ where $\Wbold = \Ibold$, the $2\times 2$ identity matrix, also providing in \cite{Vogel2013}  a fully-discrete theory for its consistent and efficient numerical realisation. In \cite{AutMorLlo18}, Dirichlet and mixed boundary conditions have been also considered.
In the case of standard Neumann b.c.\ and when $\Wbold \neq \Ibold$, the diffusivity tensor  favours anisotropic diffusion and transport, making the model adapt to directional image inpainting, see \cite{Parisotto_2019}. 

Solutions to \eqref{eq: equation directional} enjoy non-negativity and mass conservation properties. Differently with standard diffusion models, they converge to a non-constant steady state $u^*$ which \rev{can be explicitly computed channel-wise as a rescaled} version of the given image $v$ 
in terms of the ratio between the mean of $u^0$ and $v$, \rev{i.e.}
\[
u^*(\xbold) = \frac{\mu_{u^0}}{\mu_{v}} v(\xbold),\quad \xbold\in\Omega,
\]
where for any function \revbis{$g\in L^1(\Omega;\RR^3)$} we set $\mu_g := \frac{1}{|\Omega|}\int_\Omega g(\xbold)\diff\xbold$, which has to be understood channel-wise.
Model \eqref{eq: equation directional} has also  a variational counterpart, where the multiplicative nature of osmosis explicitly appears: its steady state 
is the solution of the Euler-Lagrange equation of the functional $E$ 
defined in terms of \rev{the weighted norm $\norm{\blank}_{\Wbold}:=\sqrt{\langle \blank,\, \Wbold  (\blank)\rangle}$m , see \cite[Proposition 2.3]{Parisotto_2019} as}:
\rev{
\begin{align} \label{eq: energy with directions}
E(u) 
&= 
 \int_\Omega v(\xbold)\;
 \norm{\grad \left(\frac{u(\xbold)}{v(\xbold)}\right)}^2_{\Wbold}\diff \xbold.
\end{align}
}
Images osmosis has been shown to be a very flexible tool for a variety of image processing tasks, e.g.\ image compression \cite{Weickert2013}, artefact-free shadow removal \cite{Parisotto_2019,Calatroni2017}, contrast equalisation and multi-modal image fusion in particular in the context of Cultural Heritage imaging \cite{Parisotto2018,Calatroni2018}.

From a numerical point of view, explicit, implicit \cite{Vogel2013}, and semi-implicit schemes based on the dimensional decomposition of the space-discretization operators \cite{Calatroni2017} have been considered for the efficient computation of the solution of the isotropic model. In \cite{Parisotto_2019}, suitable anisotropic stencils and efficient time-integration schemes have been introduced, providing a discrete theory for the anisotropic case. The proposed schemes exploit ideas of Lattice Basis Reduction (LBR) models proposed in \cite{Mirebeau2014} for pure diffusion processes and adapted to deal with the convection part as well. For all these schemes, \rev{consistent} mass and non-negativity \rev{preservation} properties, along with spectral convergence results, are proved. 
Note that in \cite{Parisotto_2019} the anisotropic tensor $\Wbold$ in \eqref{eq: equation directional} is computed with respect to the given image $u^0$ only, thus making the model linear. Furthermore, local coherence direction estimation is there performed once-for-all from the given image using the multi-scale tensor voting technique proposed in \cite{Moreno2012} to robustify structure tensor approaches.

\rev{In the following, we will restrict to the case $\Wbold=\Ibold$ and formulate a non-linear model where structural information is jointly updated during the fusion process.}

\subsubsection*{Non-linear models}
Non-linear models for image fusion have recently been considered in \cite{BenMolNosBurCreGilSch2017}. They are based on the computation of the (suitably defined) spectral decomposition of one-homogeneous functionals (such as the Total Variation semi-norm) and on their use as multi-scale basis. Such property is used in \cite{BenMolNosBurCreGilSch2017} to deal also with several imaging problems such as image insertion and artistic image transformation. One of the main drawbacks of this approach is the precise (and often heuristic) selection of the eigenvectors extracted from the images to fuse, which is required to combine information with enough precision. 
Moreover, in the case of face fusion, the somatic features are required to be perfectly aligned.



\section{A joint osmosis-reconstruction model}\label{sec: variational model}
We now present the osmosis-based variational model which smoothly fuses together \rev{two given images $f$ (foreground) and $b$ (background) into an image $u$}.

\rev{In order to favour the fusion between $f$ and $b$ via the osmosis energy, we will use in the following the notion of $\alpha$-map. Let $\Omega=A_f\cup A_b \cup A_\text{mix}$, with $A_f$ be a region where the foreground data is desired, ($u(\xbold) = f(\xbold)$ on $A_f$), $A_b$ where the background data is desired ($u(\xbold) = b(\xbold)$ on $A_b$), and $A_\text{mix}$ the uncertain zone, i.e.\ where $f(\xbold)$ and $b(\xbold)$ are mixed. An $\alpha$-map between $f$ and $b$ can be defined as follows:
\begin{equation}  \label{eq:alpha map}
\alpha(\xbold) = \begin{dcases}
1 &\text{if } \xbold\in A_f,\\
\alpha\in(0,1) &\text{if }\xbold\in A_\text{mix},\\
0 &\text{if }\xbold\in A_b.\\
\end{dcases}
\end{equation}
}
In this work we use \rev{user-supplied or pre-computed (via \cite{Chen2012}) $\alpha$-maps to identify the source images $f$ and $b$}, see Figure \ref{fig: dataset} for some examples. In the following section, we will make precise the three terms appearing in model \eqref{eq: joint model final}.

\subsection{Osmosis-driven fusion term}
We model the fusion process using the osmosis variational energy in \eqref{eq: energy with directions}, which is defined in terms of an additional image $v$ that has to be estimated. The idea is to use 
as $v$ an image whose associated 
drift $\grad\log v$ is related to the structural information contained both in $f$ and in $b$\revbis{.  
A possible choice consists in choosing the reference drift for $\xbold\in\Omega$ as
}
\begin{equation}
\tilde{\dbold}(\xbold)
=\grad \log\left(f(\xbold)^{\alpha(\xbold)}b(\xbold)^{1-\alpha(\xbold)}\right),
\label{eq:vector field}
\end{equation}
Note that in the case of a scalar $\alpha$-map (i.e. when $\alpha(\xbold)\equiv\alpha\in [0,1]$ for all $\xbold\in\Omega$) and by simple algebraic manipulations, this choice for $\tilde{\dbold}$ is nothing but the convex combination of the drifts associated to $f$ and $b$, respectively.
\revbis{In particular, when $\alpha\equiv 1/2$ in \eqref{eq:alpha map} then} the $\alpha$-map is simply an average of drifts corresponding to $f$ and $b$, as in \cite[Section 4.3]{Weickert2013}. However, for general $\alpha(\xbold)$, such information is locally weighted differently. This can be achieved by considering as $\alpha$-map a convolution of \eqref{eq:alpha map} with a compactly supported kernel $\omega\in C_c^\infty(\Omega)$.

In order to enforce $v$ to stay locally close to such reference image $f^\alpha b^{1-\alpha}$, we thus incorporate in our model a natural penalisation term, weighted by a parameter $\mu>0$. Combining altogether, the proposed osmosis-driven fusion term, \revbis{depending both on $u, v\in H^1(\Omega;\RR^3_+)$}, thus reads
\begingroup\makeatletter\def\f@size{8.5}\check@mathfonts
\def\maketag@@@#1{\hbox{\m@th\normalsize\normalfont#1}}
\begin{equation}\label{eq:osmo}
\mathcal{O}(u,v)= \frac{1}{2} \int_\Omega v(\xbold)\abs{\grad\left(\frac{u(\xbold)}{v(\xbold)}\right)}^2\diff \xbold +  \frac{\mu}{2} \norm{v-f^\alpha b^{1-\alpha}}_2^2.
\end{equation} 
\endgroup
This non-convex term combines osmosis with a fidelity term for $v$ which enforces local  consistence with both $f$ and $b$.


\subsection{Data fidelity and regularisation}
We combine the osmosis term \eqref{eq:osmo} with two further terms. The former acts as a fidelity term imposing $u$ to stay close to $f$ in the foreground region (i.e.\ where $\alpha(\blank)>0$):
\begin{equation}
\begin{aligned}
\mathcal{D}(u)=\frac{1}{2}\int_\Omega \alpha(\xbold) (u(\xbold)-f(\xbold)) ^2\diff\xbold.
\label{eq:data}
\end{aligned}
\end{equation}
\revbis{Note, that in order to have both \eqref{eq:osmo} and \eqref{eq:data} well-defined, we need to assume $f(\cdot)^{\alpha(\cdot)}, b(\cdot)^{1-\alpha(\cdot)}\in L^2(\Omega;\RR^3_+)$ and $f\in  L^2(\Omega;\RR^3_+)$.}
We remark that although the quadratic term in \eqref{eq:osmo} also plays the role of a data fidelity on $v$ and, as such, it could be combined with the term in \eqref{eq:data}, we nonetheless decided to leave it in the definition of  $\mathcal{O}$ to underline the importance of its closeness to the image $f^\alpha b^{1-\alpha}$ \rev{for favouring} the osmosis fusion driven by $v$. \rev{In the following numerical section, we thus respect such a decomposition}.

\smallskip

As we are going to show in our following numerical experiments, for many real-world applications the combination of $\mathcal{O}$ and $\mathcal{D}$ already produces satisfying results. However, whenever prior information is available on the image $v$, the model can even be improved by introducing a further regularisation term acting on $v$ which will be denoted by $\mathcal{R}(v)$. \rev{A possible choice for this term is} the $L^1$ norm of the image gradient, so as to promote gradient sparsity 
\begin{equation}  \label{eq: regularisation R}
\Rcal(v) = \int_\Omega \abs{\grad v(\xbold)}\diff\xbold.
\end{equation}
This choice corresponds to a piece-wise fusion of the image information in $v$, in contrast to the smoothing introduced by the $\alpha$-map, so as to fuse only the relevant features from the foreground $f$ and  background $b$, e.g.\ in text fusion applications as in Figure \ref{fig: Verona manuscript}.  

\smallskip

These three terms are merged in the final proposed model \eqref{eq: joint model final}. 
%
which we recall here for better readability
\[
\min_{u,v} ~ \Ecal(u,v):=\Ocal(u,v)+\gamma\Dcal(u)+\eta\Rcal(v).
\]
By varying the  weights \revbis{$(\mu,\gamma,\eta)$}, we observe different behaviors on the final images depending also on the initial condition considered. This is due to the non-convexity of the model whose careful analysis, \revbis{e.g.\ of the existence of local minimizers,} is left for future research. Empirical considerations on these aspects are discussed in Section \ref{sec: applications}.

\section{Numerical solution}\label{sec: numerical details}

The energy in \eqref{eq: joint model final} is composed of three terms: the osmosis term $\Ocal(u,v)$ defined in \eqref{eq:osmo}, coupling two smooth terms \rev{in a jointly non-convex way},
the convex and smooth data term $\Dcal(u)$ \eqref{eq:data} and the regularisation function $\Rcal(v)$\rev{, which can be either smooth (e.g.\ Tikhonov-type) or not (cf. \eqref{eq: regularisation R})}.

In order to exploit \rev{the smoothness and convexity properties} of these terms to compute a local minimum of problem \eqref{eq: joint model final}, we can apply the \emph{Inertial Proximal Algorithm for Non-Convex Optimization} (iPiano) algorithm \cite{Ochs}. \rev{Denoting for $k\geq 0$ by $u^{k}$ and $v^{k}$ the $k-$th iterates,} the corresponding iterative scheme reads \rev{in a continuum setting}:
\begingroup\makeatletter\def\f@size{8.5}\check@mathfonts
\def\maketag@@@#1{\hbox{\m@th\normalsize\normalfont#1}}
\begin{equation}\label{eq:ipiano}
\begin{dcases}
u^{k+1}= \prox_{\zeta_1^k\gamma\mathcal{D}} (u^k-\zeta_1^k \partial_u \mathcal{O}(u^k,v^{k}) +\beta_1 (u^k - u^{k-1})),\\
v^{k+1}= \prox_{\zeta_2^k\eta\mathcal{R}} (v^k -\zeta_2^k \partial_v \mathcal{O}(u^{k},v^{k}) + \beta_2(v^{k}-v^{k-1})),
\end{dcases}
\end{equation}
\endgroup
 for some initial guesses $u^0, v^0$, inertial scalar steps $\beta_1,\beta_2\rev{\in[0,1)}$ and time-step sequences $\zeta_1^k,\zeta_2^k$, automatically computed via backtracking \rev{of} the Lipschitz constant of $\partial_u \Ocal(u,v)$ and $\partial_v\Ocal(u,v)$ via estimates $L_{1,k}$ and $L_{2,k}$, respectively, via
 \[
 \zeta_1^k < 2(1-\beta_1)/L_{1,k}\quad\text{and}\quad\zeta_2^k < 2(1-\beta_2)/L_{2,k}.
 \]
Following \cite{Ochs}, $L_{1,k}\in\{L_{1,k-1}, \lambda L_{1,k-1}, \lambda^2 L_{1,k-1},\dots\}$\rev{, where $\lambda>1$,} 
 is the minimal value satisfying, \rev{for $p_1=u^{k+1}$ and $p_2=v^{k+1}$ , the conditions: }
\begingroup\makeatletter\def\f@size{8.5}\check@mathfonts
    \begin{equation*}
    \rev{
    \begin{aligned}
    \Ocal(p_1,v^k) &\leq
    \Ocal(u^k,v^k) + \langle \partial_u \Ocal(u^k,v^k),\,p_1-u^{k}\rangle+\frac{L_{1,k}}{2}\norm{p_1-u^{k}}_2^2,\\
    \Ocal(u^k,p_2) &\leq
    \Ocal(u^k,v^k) + \langle \partial_u \Ocal(u^k,v^k),\,p_2-v^{k}\rangle+\frac{L_{2,k}}{2}\norm{p_2-v^{k}}_2^2.
    \end{aligned}
    }
    \end{equation*}
\endgroup
 
The algorithm requires the computation of the \revbis{weak} derivative of the smooth term $\mathcal{O}(u,v)$ and the proximal operators of the other two terms. 
We now detail these computations, while the whole process \rev{and the default values of iPiano parameters are} presented in Algorithm \ref{alg: joint TV-osmosis block-coordinate iPiano scheme}. 

\subsubsection{Derivatives of $\mathcal{O}(u,v)$}
The \revbis{weak} gradient with respect to the variable $u$ is computed as
\begin{equation}
\revbis{\partial_u \mathcal{O}(u,v) = 
-\left( \lap u - \div(\dbold u)\right)(v^{-1})}
\label{eq: grad u g2}
\end{equation}
where $\dbold:=\grad\log v$, $\grad^*=-\div$, $\grad^*\grad=-\Delta$ and where we assumed Neumann boundary conditions $\langle\grad u - \dbold u, \nbold\rangle=0$.
The gradient with respect to $v$ is (see Appendix \ref{appendix}):
\begingroup\makeatletter\def\f@size{8.5}\check@mathfonts
\def\maketag@@@#1{\hbox{\m@th\normalsize\normalfont#1}}
\begin{equation}
\begin{aligned}
\partial_v \mathcal{O}(u,v) 
=& 
-\frac{1}{2v^2}\abs{\grad u}^2
+\frac{2u}{v^3} \grad u\cdot \grad v
+ \div\left( \frac{u}{v^2} \grad u \right)\\
&- \frac{3u^2}{2v^4} \abs{\grad v}^2
- 2v \div\left( \frac{u^2}{v^3}\grad v \right)\\
&+ \mu (v-f^\alpha b^{1-\alpha}).
\end{aligned}
\label{eq: grad v g2}
\end{equation}
\endgroup

\subsubsection{Proximal operator of $\mathcal{D}(u)$}
The proximal step is simply
\begingroup\makeatletter\def\f@size{8.5}\check@mathfonts
\def\maketag@@@#1{\hbox{\m@th\normalsize\normalfont#1}}
\begin{equation}
\begin{aligned}
\prox_{\zeta_1^k\gamma \mathcal{D}} (u^{k}_\diamond)&\hspace{-0.05cm}=\hspace{-0.05cm}
\argmin_{u} \frac{\gamma}{2}\norm{\sqrt{\alpha}(u-f)}_2^2 \hspace{-0.05cm}+\hspace{-0.05cm} \frac{1}{2\zeta_1^k}\norm{u-u^{k}_\diamond}_2^2\\
&\hspace{-0.05cm}=\hspace{-0.05cm}
\left(\gamma \alpha + \frac{1}{\zeta_1^k}\right)^{-1} \left( \gamma \alpha f + \frac{1}{\zeta_1^k} u^{k}_\diamond\right).
\label{eq: proximal ipiano u}
\end{aligned}
\end{equation}
\endgroup
\subsubsection{Proximal operator of $\mathcal{R}(v)$}
If $\mathcal{R}$ is smooth (for instance, if Tikhonov regularisation $\mathcal{R}(v)=\frac12||\grad v||^2$ is used),  explicit gradient descent can be performed instead of computing an implicit proximal point.
 The current step can thus be integrated with $\partial_v\mathcal{O}$ so that the full update on $v$ in \eqref{eq:ipiano} becomes:
$$v^{k+1}= v^k - \zeta_2^k (\partial_v \mathcal{O}(u^{k},v^{k})+\partial_v \mathcal{R}(v^{k}) ) + \beta_2(v^{k}-v^{k-1}).$$

However, implicit schemes need to be considered when $\mathcal{R}$ is not smooth, e.g.\ when $\mathcal{R}$ is chosen as in \eqref{eq: regularisation R}.
A standard way-around to improve the computational efficiency for such calculation consists in substituting the TV-type term with with its Huber regularisation $\mathcal{H}_\varepsilon$ depending on a small parameter $0<\varepsilon\ll 1$, see, e.g., \cite{ChaPoc2011} for details. This promotes quadratic regularisation for homogeneous areas where $\norm{\grad v}_1\leq \varepsilon$, and $L^1$ gradient regularisation wherever $\norm{\grad v}_1> \varepsilon$.
The proximal step for the Huberised TV is thus computed \rev{by solving}
\begingroup\makeatletter\def\f@size{9.5}\check@mathfonts
\def\maketag@@@#1{\hbox{\m@th\normalsize\normalfont#1}}
\begin{equation}
\prox_{\zeta_2^k \eta\mathcal{H}_\varepsilon}\left( v^{k}_\diamond \right)
= 
\argmin_v ~\eta \mathcal{H}_\varepsilon(\grad v) + \frac{1}{2\zeta_2^k}\norm{v-v^{k}_\diamond}_2^2.
\label{eq: implicit proximal ipiano v}
\end{equation}
\endgroup
Following \cite{ChaPoc2011}, \revbis{the saddle point problem from \eqref{eq: implicit proximal ipiano v} is}\footnote{where we denote by $p$ the minimization variable instead of $v$ in order to avoid confusion between these inner iterations and the main ones.}
\begingroup\makeatletter\def\f@size{9.5}\check@mathfonts
\def\maketag@@@#1{\hbox{\m@th\normalsize\normalfont#1}}
\begin{equation}
\label{eq:innerPD}
\min_p \max_{\ybold} ~
-\mathcal{H}^*_\varepsilon(\ybold)+\underbrace{ \langle\grad p,\, \ybold\rangle+\frac{1}{2\zeta_2^k}\norm{p-v^{k}_\diamond}_2^2}_{s(p)}.
\end{equation}
\endgroup
where, with a little abuse of notation, we have defined
$
\mathcal{H}_\varepsilon^*(\ybold):=\left(\eta \mathcal{H}_\varepsilon\right)^*(\ybold)= \frac{\varepsilon}{2\eta}\norm{\ybold}^2 + \delta_{\norm{\blank}_{2,\infty}\leq \eta}(\ybold)
$.
Since $s(p)$ is strongly convex and taking $\tau_0,\sigma_0>0$ such that $\tau_0\sigma_0 L^2\leq 1$, where $L^2=\norm{\grad}^2$ is the squared operator norm, then a forward discretization of \rev{the operator} $\grad$ entails $\norm{\grad}^2\leq 8$. 

\noindent
The accelerated primal-dual algorithm  \cite{ChaPoc2011} solving \eqref{eq:innerPD}  reads:
\begin{equation}
\begin{aligned}
\ybold^{t+1} &= \prox_{\sigma_t \mathcal{H}_\varepsilon^*}(\ybold^t+\sigma_t \grad \overline{p}^t)\\
p^{t+1} &= \prox_{\tau_t s} (p^t+\tau_t \div \ybold^{t+1})\\
\omega_t &= (1+2\hat{\gamma} \tau_t)^{-0.5},\, \tau_{t+1}=\omega_t \tau_t,\, \sigma_{t+1}=\sigma_t/\omega_t \\
\overline{p}^{t+1} &= p^{t+1} + \omega_t (p^{t+1}-p^{t}),
\end{aligned}
\label{eq: nested pd}
\end{equation}
with proximal maps
\begingroup\makeatletter\def\f@size{8.5}\check@mathfonts
\def\maketag@@@#1{\hbox{\m@th\normalsize\normalfont#1}}
\[
\begin{aligned}
\prox_{\sigma \mathcal{H}_\varepsilon^*}(\ybold) &= \frac{\ybold}{1+\sigma\varepsilon}\left(\max\left\{1,\eta^{-1}\norm{\frac{\ybold}{1+\sigma\varepsilon}}_2\right\}\right)^{-1},&\\
\prox_{\tau s}(p) &= \left(\frac{1}{\zeta_2^k} + \frac{1}{\tau}\right)^{-1}\left( \frac{{v^{k}_\diamond}}{\zeta_2^k} +\frac{p}{\tau}\right).&
\end{aligned}
\]
\endgroup
Algorithm \eqref{eq: nested pd} converges to a saddle point $(p^*,\ybold^*)$ of \eqref{eq:innerPD}, so we deduce $\prox_{\zeta_2^k \eta\mathcal{H}_\varepsilon}\left( v^{k}_\diamond \right)=p^*$.

\begin{algorithm*}[tb]\noindent 
\SetAlgoLined\footnotesize
\caption{Block-coordinate iterative scheme for the joint osmosis model \eqref{eq: joint model final}}
\label{alg: joint TV-osmosis block-coordinate iPiano scheme}
\SetKwData{maxiter}{maxiter\_model}
\SetKwData{maxiterip}{maxiter\_iPiano}
\SetKwData{gapu}{gap\_u}
\SetKwData{gapv}{gap\_v}
\SetKwData{exitflag}{exit\_flag}
\SetKwData{tol}{tol}
\SetKwData{prox}{prox}
\SetKwData{colourchannel}{color channel}
\SetKwInOut{Input}{Input}
\SetKwInOut{Output}{Output}
\SetKwInOut{Parameters}{Parameters}
\SetKwProg{Fn}{Function}{:}{}
\SetKwFunction{primaldual}{primal\_dual}
\Input{foreground image $f$, background image $b$. $f,b:\Omega\to[0,255]$. $\alpha$-map $\alpha:\Omega\to[0,1]$;}
\Output{the fused image $u$;}
\Parameters{for the algorithm: tolerance (\tol) and maximum number of iterations (\maxiter); 
for the model \eqref{eq: joint model final}: weights $(\eta,\mu,\gamma)$;\\
for iPiano: \rev{$\zeta_{1,0}= 0.99(1-2\beta_1)/L_1, \zeta_{2,0}= 0.99(1-2\beta_2)/L_2$ with $\beta_1=\beta_2=0.4$ and $L_{1,0}= L_{2,0}=1$}, \maxiterip;
}
\BlankLine
\Fn{block\_coordinate\_osmosis}{
\BlankLine
$u^0 = f$ (other choices in Figure \ref{fig: puppets}) and $v^0 = f^\alpha b^{1-\alpha}$\tcp*{Initialisation}
\BlankLine
\For{$k=0,\dots,\maxiter$}{
\BlankLine
\tcp{Compute the explicit iPiano steps}
$u_\diamond^{k} = u^k - \zeta_{1,k} \partial_u \mathcal{O}(u^k,v^k) + \beta_1 (u^k - u^{k-1})$\tcp*{with $\partial_u \mathcal{O}$ as in \eqref{eq: grad u g2}}
$v_\diamond^{k} = v^k - \zeta_{2,k} \partial_v \mathcal{O}(u^k,v^k) + \beta_2 (v^k - v^{k-1})$\tcp*{with $\partial_v \mathcal{O}$ as in \eqref{eq: grad v g2}}
\BlankLine
\tcp{Compute the implicit (proximal) steps of iPiano}
\For{$t=0,\dots,\maxiterip$}{
$p_1 = \prox_{\zeta_{1,k} \gamma\mathcal{D}}\left( u_\diamond^{k}\right)$\tcp*{computed via \eqref{eq: proximal ipiano u}}
$p_2 = \prox_{\zeta_{2,k}\eta\mathcal{R}}\left( v_\diamond^{k} \right)$\tcp*{via the primal-dual Algorithm in \eqref{eq: nested pd}}
\gapu = $\Ocal(p_1,v^k) - \Ocal(u^k,v^k) - \langle \partial_u \Ocal(u^k,v^k),(p_1-u^k)\rangle - (L_{1,k}/2)\norm{p_1-u^k}_2^2 $\;
\gapv = $\Ocal(u^k,p_2) - \Ocal(u^k,v^k) - \langle \partial_v \Ocal(u^k,v^k),(p_2-v^k)\rangle - (L_{2,k}/2)\norm{p_2-v^k}_2^2 $\;
\BlankLine
\If{
$\gapu < 0$ and $\gapv < 0$}
{
$u^{k+1} = p_1$; $v^{k+1} = p_2$; \Break \tcp*{accept and update the variables}
}
Update $\zeta_{1,k},\zeta_{2,k},L_{1,k},L_{2,k}$\tcp*{according to the standard iPiano rules}
}
\BlankLine
\IfThen{
$\abs{\Ecal(u^{k+1},v^{k+1})-\Ecal(u^{k},v^{k})}/\abs{\Ecal(u^{k+1},v^{k+1})}<\tol$
}{
\Break}\tcp*{Exit condition from the main loop}
}
\BlankLine
}
\Return
\end{algorithm*}

\paragraph*{Parameters} In the next section we describe different applications of Algorithm \ref{alg: joint TV-osmosis block-coordinate iPiano scheme} to imaging problems and test its sensitivity to the weights $(\eta,\mu,\gamma)$, with \rev{$\varepsilon$-Huberised smoothing of the $L^1$ gradient norm with $\varepsilon=0.05$}. For the iPiano algorithm we use the following parameters: the exit condition for the iPiano (with backtracking) scheme is chosen to satisfy either the relative error on the energy (\texttt{1e-6}) or the maximum number of iterations (10000), with at least few iterations performed so as to assure the inertial contribution; the exit condition for the nested primal-dual problem \eqref{eq: nested pd} is set to satisfy either the relative error on the primal-dual gap (\texttt{1e-4}) or the maximum number of iterations (10000).

\section{Applications}\label{sec: applications}
\begin{figure*}[tbh]
    \centering
    \begin{subfigure}[b]{0.14\textwidth}\centering
    \captionsetup{justification=centering}
    \includegraphics[width=1\textwidth]{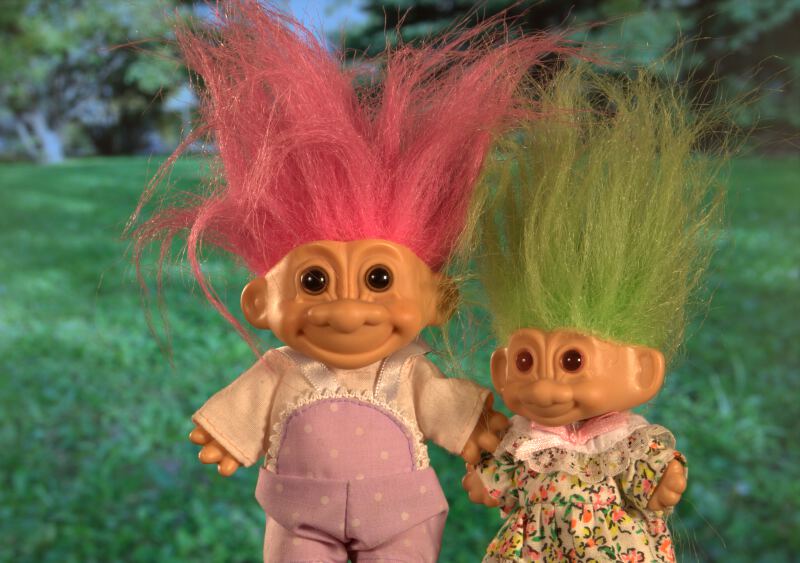}
    \caption{Foreground $f$\\(800$\times$563 px)}
    \label{fig: puppets1}
    \end{subfigure}
    \,
    \begin{subfigure}[b]{0.14\textwidth}\centering
    \captionsetup{justification=centering}
    \includegraphics[width=1\textwidth]{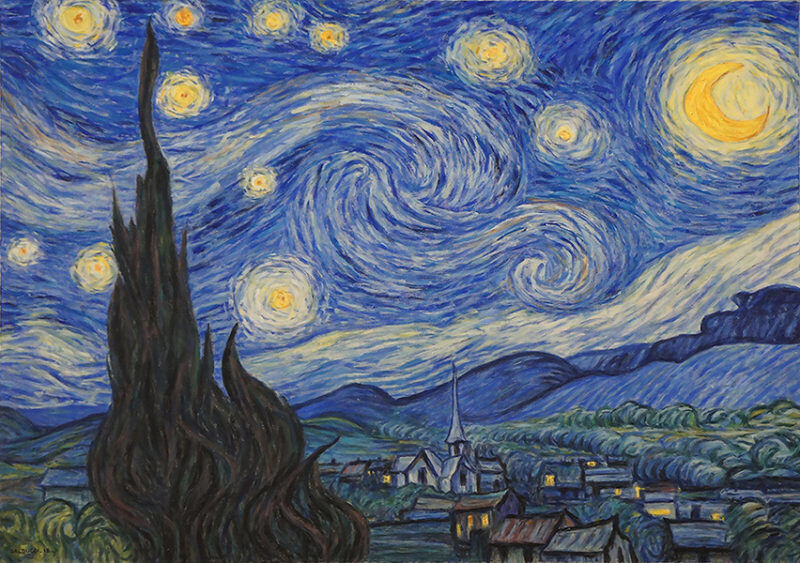}
    \caption{Background $b$\\(800$\times$563 px)}
    \label{fig: vangogh}
    \end{subfigure}
    \,
    \begin{subfigure}[b]{0.14\textwidth}\centering
    \captionsetup{justification=centering}
    \includegraphics[width=1\textwidth]{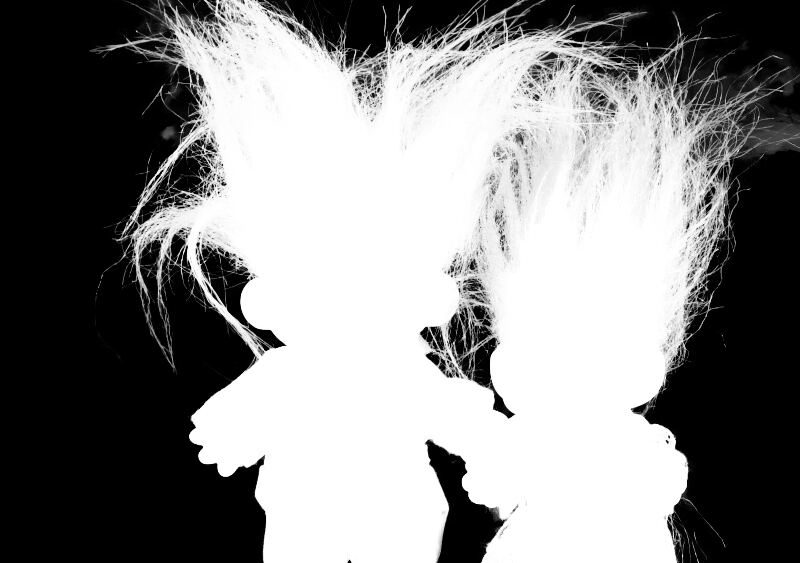}
    \caption{$\alpha$-map\\for images \ref{fig: puppets1} and \ref{fig: vangogh}.}
    \label{fig: alphamap1}
    \end{subfigure}
    \hfill
    \begin{subfigure}[t]{0.14\textwidth}\centering
    \captionsetup{justification=centering}
    \includegraphics[width=0.83\textwidth]{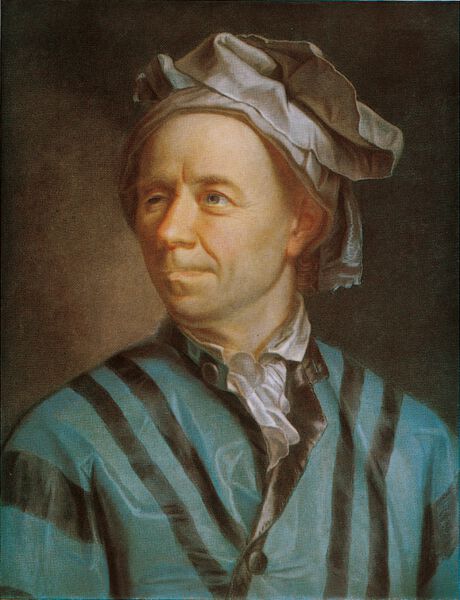}
    \caption{Foreground $f$\\ (460$\times$600 px)}
    \label{fig: Euler}
    \end{subfigure}
    \,
    \begin{subfigure}[t]{0.14\textwidth}\centering
    \captionsetup{justification=centering}
    \includegraphics[width=0.83\textwidth]{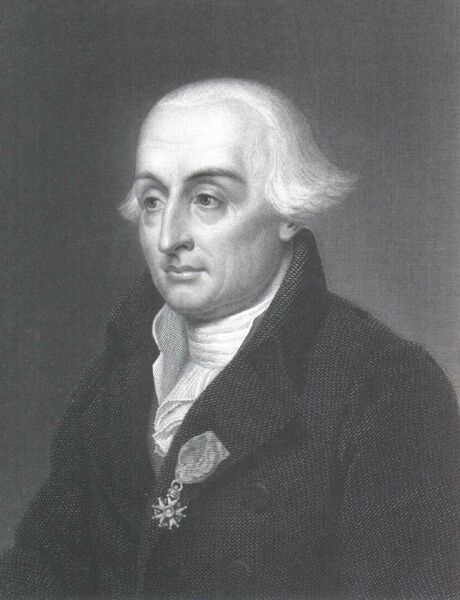}
    \caption{Background $b$\\ (460$\times$600 px)}
    \label{fig: Lagrange}
    \end{subfigure}
    \,
    \begin{subfigure}[t]{0.14\textwidth}\centering
    \captionsetup{justification=centering}
    \includegraphics[width=0.83\textwidth]{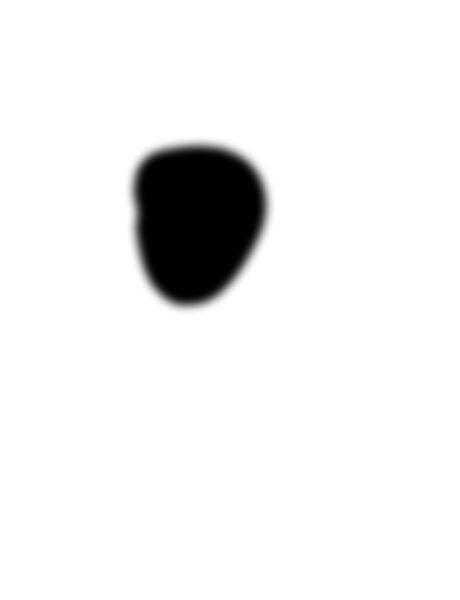}
    \caption{$\alpha$-map (blurred)\\for images  \ref{fig: Euler} and \ref{fig: Lagrange}.}
     \label{fig: alphamap2}
    \end{subfigure}
    \\ 
    \begin{subfigure}[t]{0.14\textwidth}\centering
    \captionsetup{justification=centering}
    \includegraphics[width=0.98\textwidth]{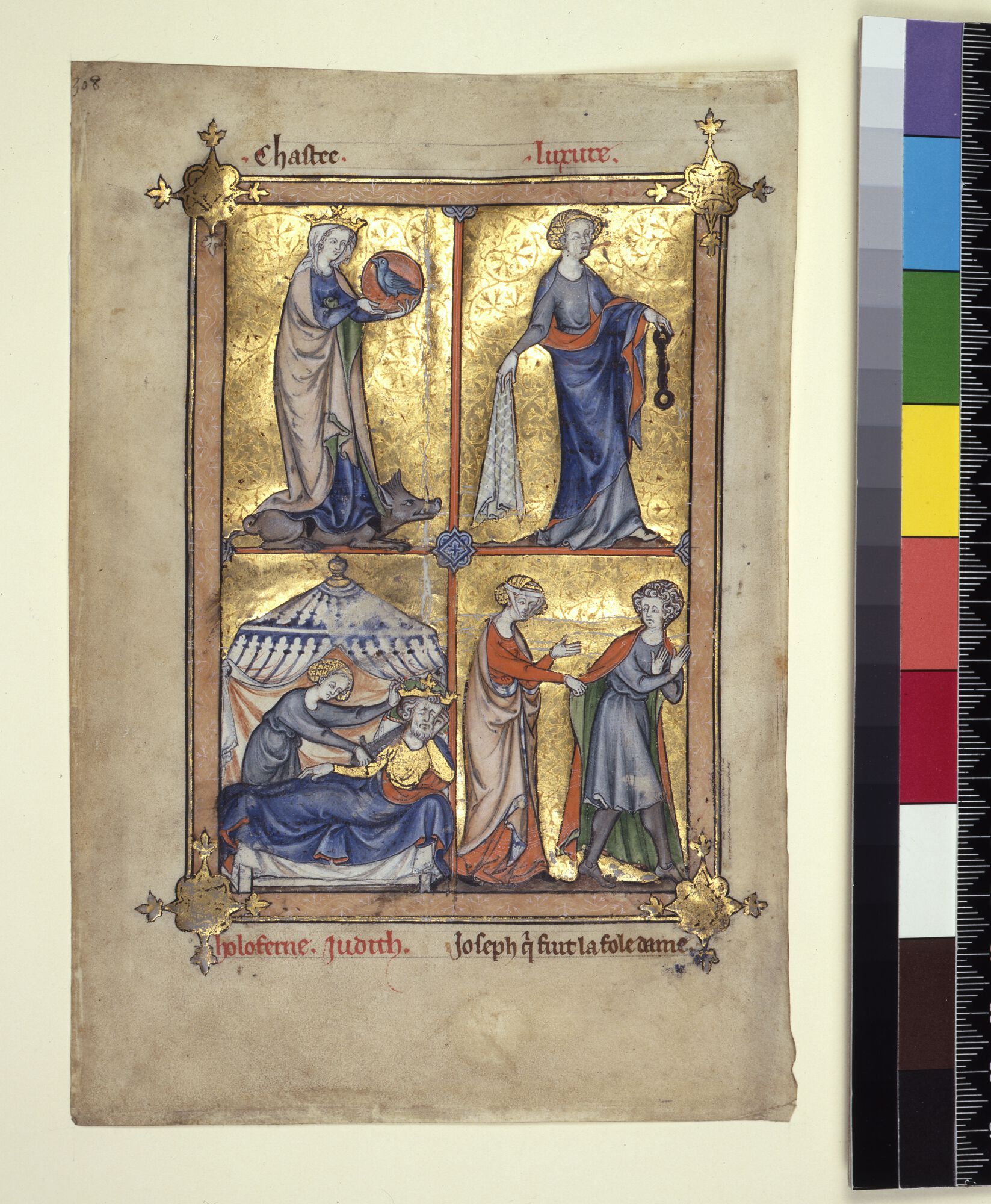}
    \caption{Foreground $f$\\($1647\times2000$ px)}
    \label{fig: fitzwilliam VIS}
    \end{subfigure}
    \hfill
    \begin{subfigure}[t]{0.14\textwidth}\centering
    \captionsetup{justification=centering}
    \includegraphics[width=0.98\textwidth]{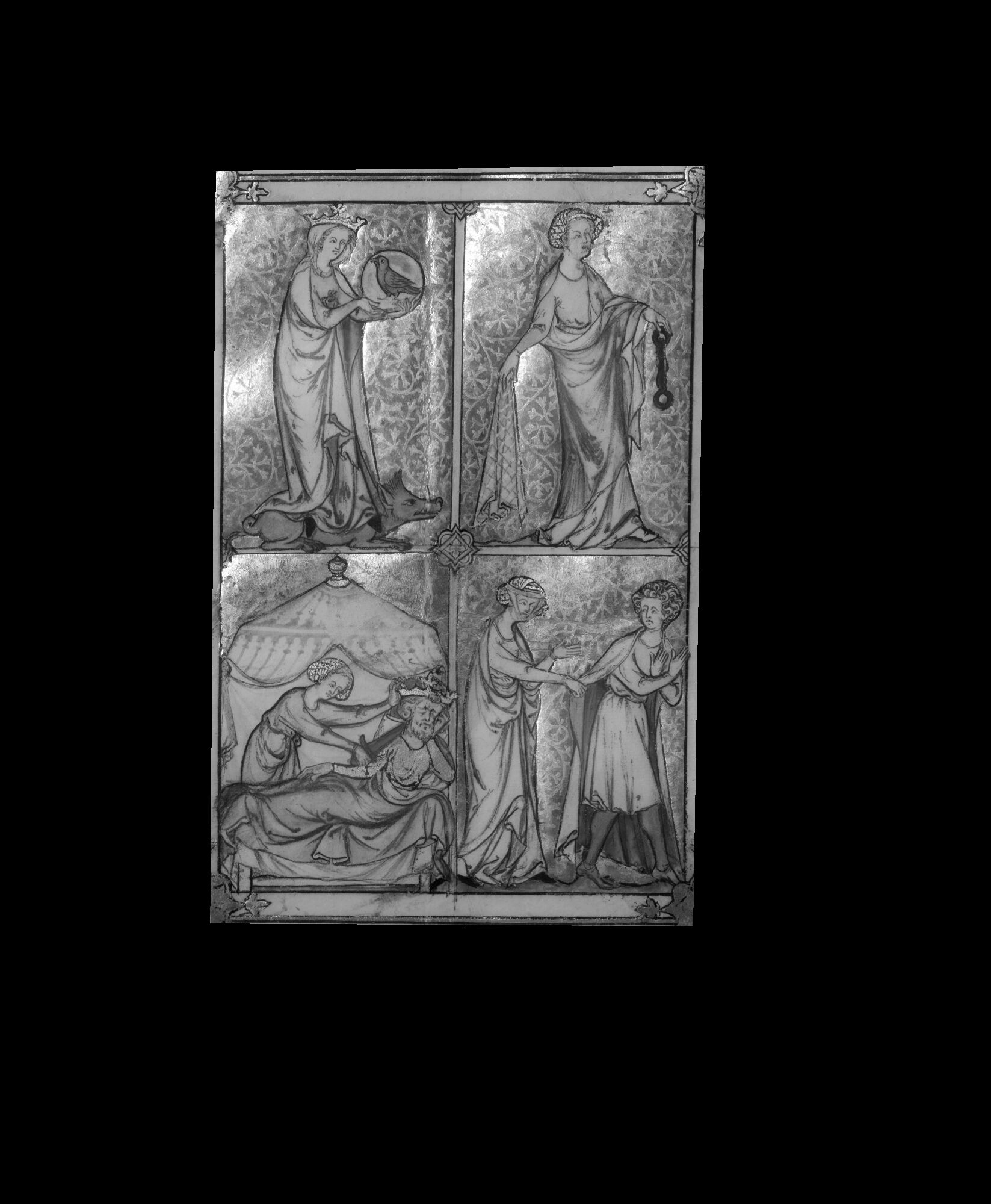}
    \caption{Background $b$\\(SIFT-registered)}
    \label{fig: fitzwilliam IR}
    \end{subfigure}
    \hfill
    \begin{subfigure}[t]{0.14\textwidth}\centering
    \captionsetup{justification=centering}
    \includegraphics[width=0.98\textwidth]{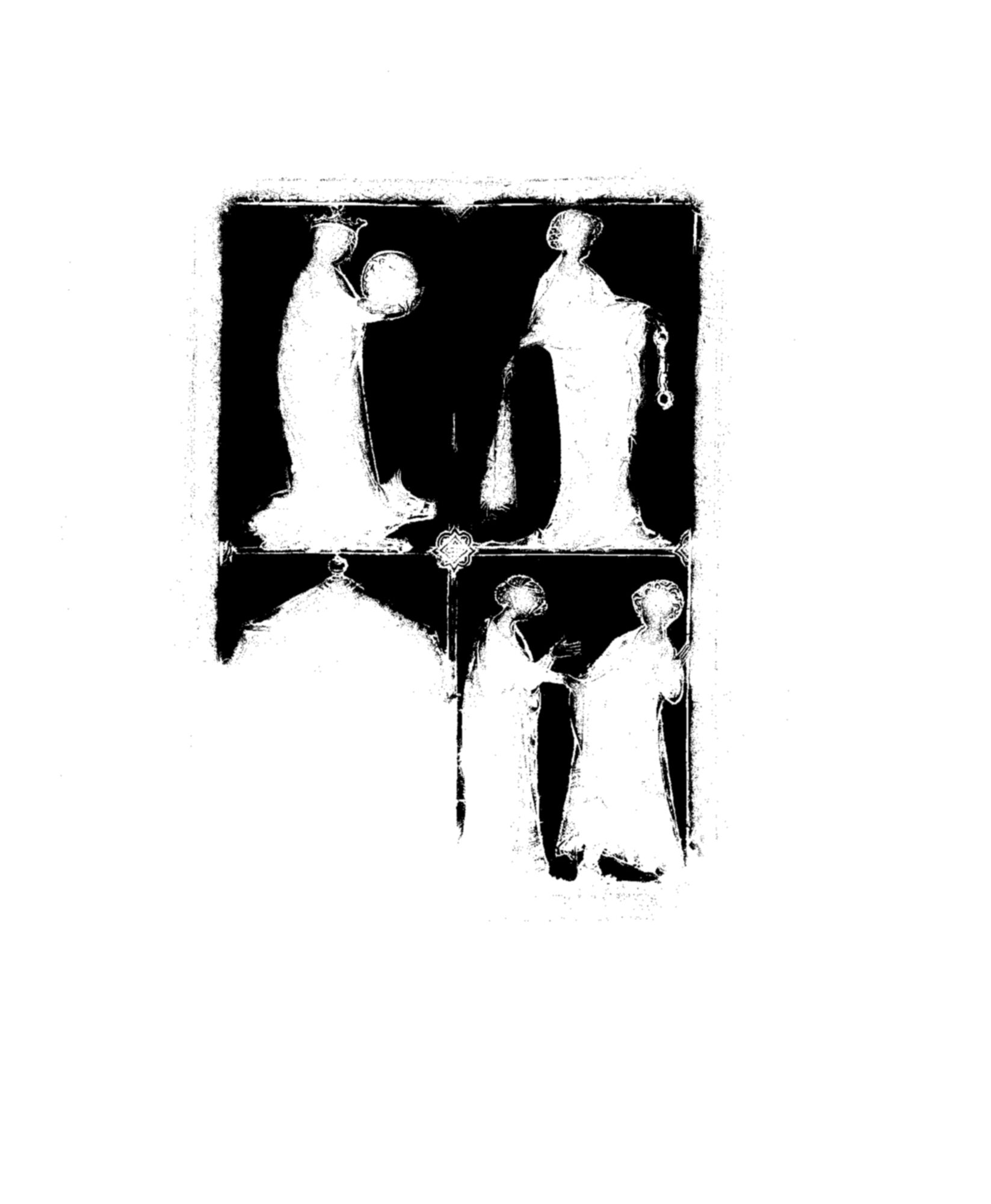}
    \caption{$\alpha$-map\\for images \ref{fig: fitzwilliam VIS} and \ref{fig: fitzwilliam IR}.}
      \label{fig: alphamap3}
    \end{subfigure}
    \hfill
    \begin{subfigure}[t]{0.14\textwidth}\centering
    \captionsetup{justification=centering}
    \includegraphics[width=0.83\textwidth,trim=0.0cm 1cm 0cm 1cm,clip=true]{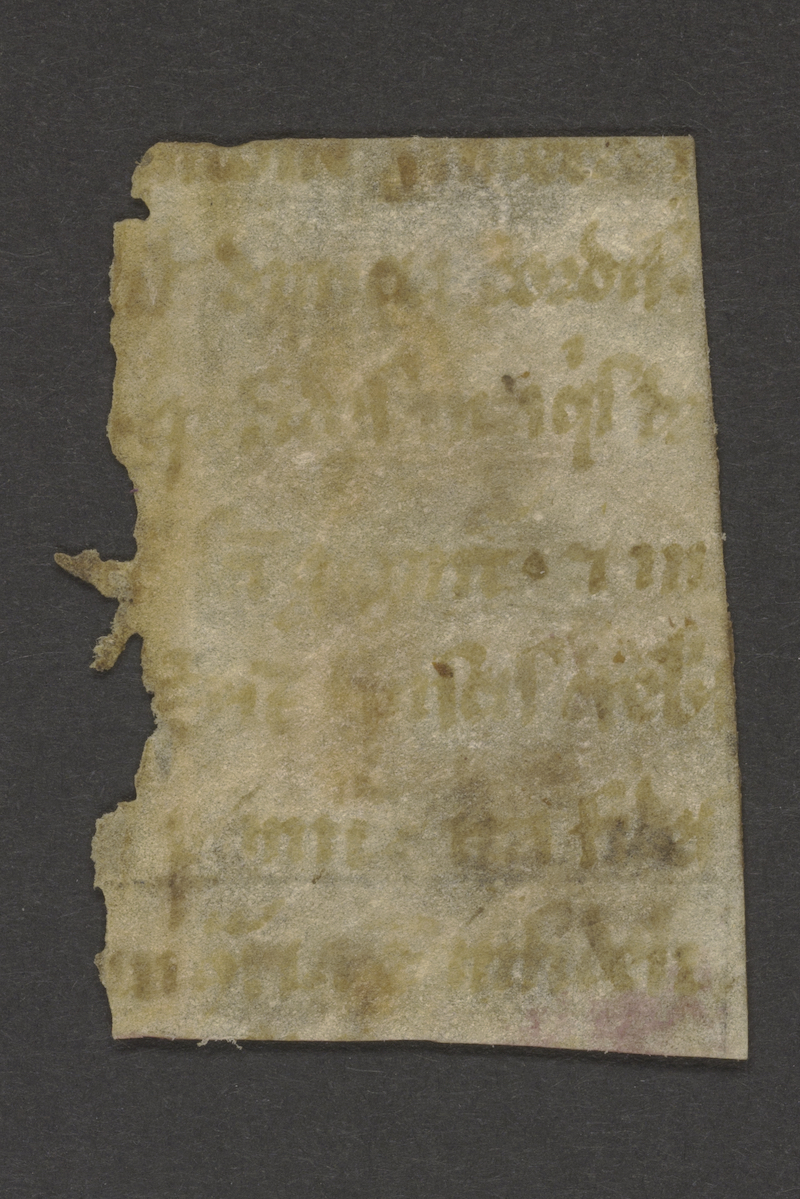}
    \caption{Foreground $f$\\($953\times636$ px)}
    \label{fig: Verona VIS}
    \end{subfigure}
    \,
    \begin{subfigure}[t]{0.14\textwidth}\centering
    \captionsetup{justification=centering}
    \includegraphics[width=0.83\textwidth,trim=0.0cm 1cm 0cm 1cm,clip=true]{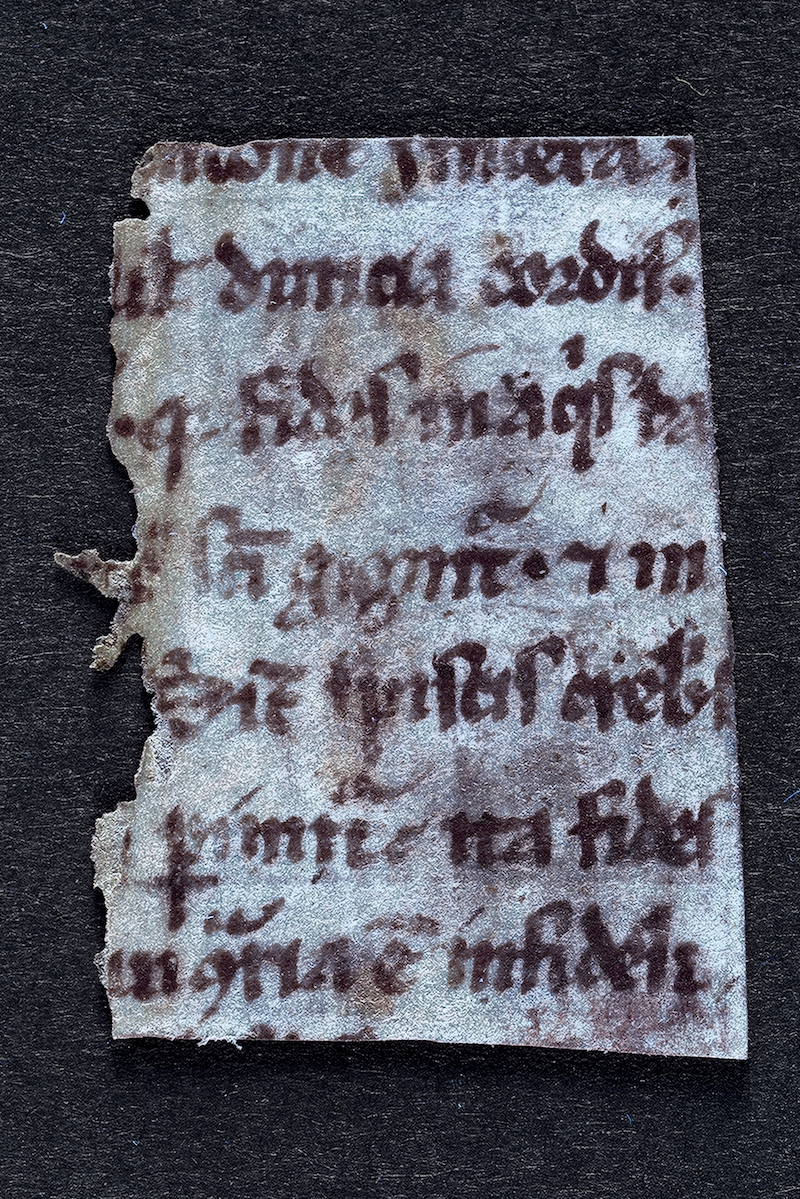}
    \caption{Background $b$\\($953\times636$ px)}
    \label{fig: Verona IR}
    \end{subfigure}
    \,
    \begin{subfigure}[t]{0.14\textwidth}\centering
    \captionsetup{justification=centering}
    \includegraphics[width=0.83\textwidth,trim=0.0cm 1cm 0cm 1cm,clip=true]{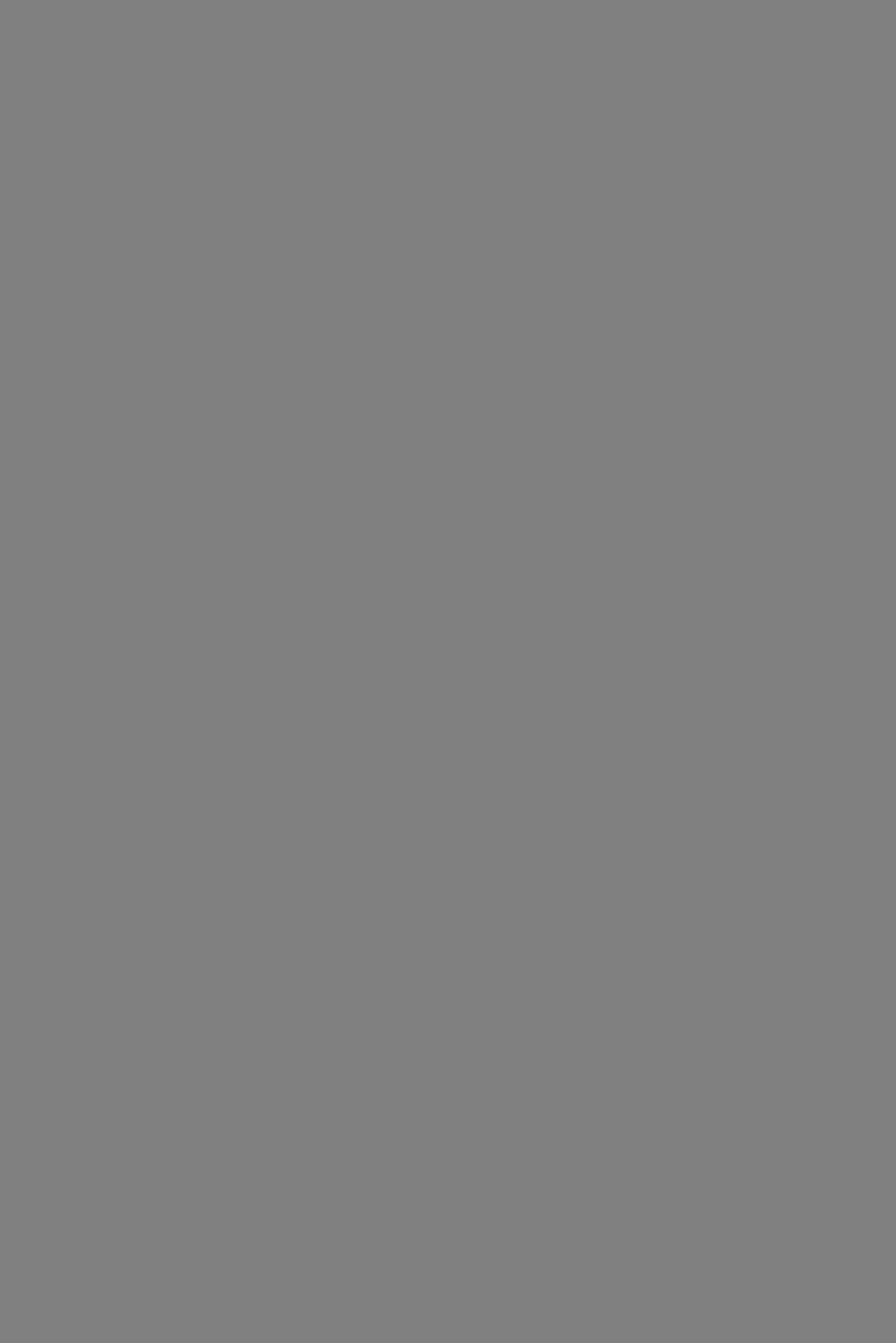}
    \caption{$\alpha$-map (= 0.5 const.)\\for images \ref{fig: Verona VIS} and \ref{fig: Verona IR}.}
      \label{fig: alphamap4}
    \end{subfigure}
    \caption{Our dataset for the data fusion applications described in this paper. Fig.\ \ref{fig: puppets1} from \url{https://github.com/dingzeyuli/knn-matting}; Fig.\ \ref{fig: vangogh} \emph{The Starry Night}, by V.\ Van Gogh, 1889 (CC-PD-Mark 1.0); Fig.\ \ref{fig: Euler}: \emph{Portrait of Leonhard Euler (1707-1783)} by J.E.\ Handmann, Kunstmuseum Basel (CC-PD-Mark 1.0
    );
    Fig.\ \ref{fig: Lagrange} \emph{Portrait of Joseph Louis Lagrange} (CC-PD-Mark 1.0);
    Figs.\ \ref{fig: fitzwilliam VIS} and \ref{fig: fitzwilliam IR}: visible (RGB) and infrared images of \emph{La Somme le roi} by Master Honor\'e's, MS 368 (by permission of the Syndics of the \emph{Fitzwilliam Museum}, Cambridge, UK); Figs.\ \ref{fig: Verona VIS} and \ref{fig: Verona IR}: visible (RGB) and infrared images of a manuscript with text (courtesy of \emph{Biblioteca Capitolare}, Verona, Italy); Figs.\ \ref{fig: alphamap1}, \ref{fig: alphamap2}, \ref{fig: alphamap3} and \ref{fig: alphamap4}: $\alpha$-maps.
    }
    \label{fig: dataset}
\end{figure*}

We apply the model \eqref{eq: joint model final} to fuse data from two given images $f$ and $b$ for different applications\footnote{\revbis{MATLAB code: \url{git.io/simoneparisotto},  \url{codeocean.com/capsule/2404871}.}}
The model requests as a further input an $\alpha$-map as defined in \eqref{eq:alpha map} indicating from which of the two images the structural information comes from. \rev{Recalling the decomposition into $A_f$, $A_b$ and $A_\text{mix}$, we notice that when no uncertainty is modelled, then $A_\text{mix}=\emptyset$ and $\alpha(\blank)$} is binary, i.e. $\alpha(\xbold)\in\left\{0,1\right\}$. 
Alternatively, $\alpha$ can be smoothed by means of a Gaussian convolution \rev{so as to model uncertainty and favour a smooth transition on $A_\text{mix}\neq \emptyset$}. In the following experiments, we use the $\alpha$-matting estimator proposed in \cite{Chen2012}.  In some cases, we had to align the images $f$ and $b$, especially for matching the boundaries of the gluing areas $A_\text{mix}$ where $\alpha\in(0,1)$. For this reason, we sometimes need to perform a pre-processing registration step using SIFT features \cite{SIFT}: this is the case for images (g)--(h) but not for (a)--(b) and (d)--(e) since the fusion zone is naturally consistent (e.g.\ the skin of the faces). \rev{More refined registration approaches such as the one proposed in \cite{Feng2019} can also be used. However, registration via SIFT keypoints worked well enough for our purposes.}
We showcase the images $f$, $b$ and $\alpha$ used in the different experiments presented in Figure \ref{fig: dataset}.

\subsection{Influence of the weights} 
\begin{figure*}
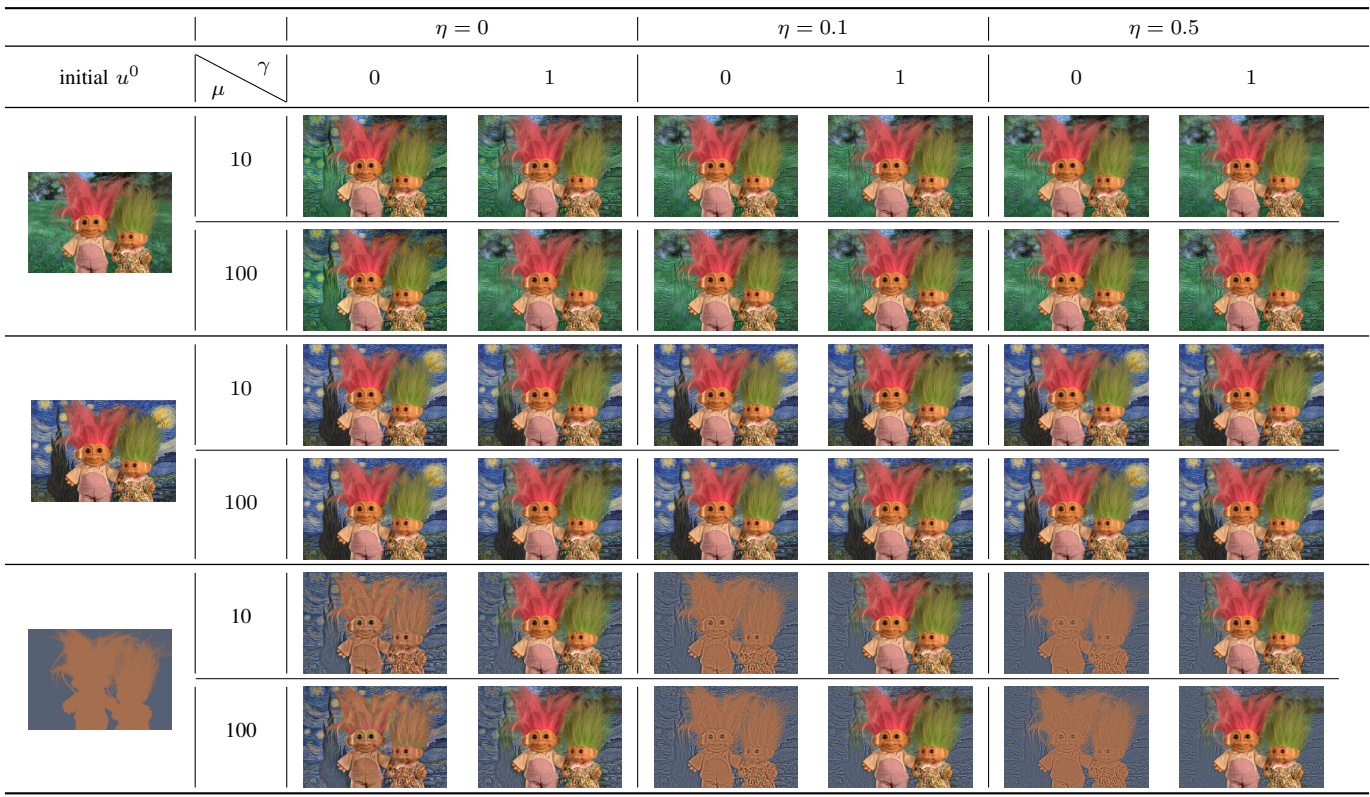
\footnotesize
\noindent
\begin{tabularx}{\textwidth}{c|c|cc|cc|cc}
\toprule
& &
\multicolumn{2}{c|}{$\eta=0$}& 
\multicolumn{2}{c|}{$\eta=0.1$}& 
\multicolumn{2}{c}{$\eta=0.5$}\\
\midrule
initial $u^0$ & \diagbox{$\mu$}{$\gamma$} & $0$ & $1$ & $0$ & $1$ & $0$ & $1$\\
\midrule
\multirow{2}{*}[-0.3cm]{
\adjincludegraphics[valign=M,width=0.105\textwidth]{{./results_puppets/1_foreground}.jpg}
}
& $10$ 
& 
\adjincludegraphics[valign=M,width=0.105\textwidth]{{./results_puppets/1_output_u_init0_alphablur0_nu0_eta1_mu10_gamma0_eps0.05_time956.1395}.jpg}
&
\adjincludegraphics[valign=M,width=0.105\textwidth]{{./results_puppets/1_output_u_init0_alphablur0_nu0_eta1_mu10_gamma1_eps0.05_time459.8769}.jpg}
&
\adjincludegraphics[valign=M,width=0.105\textwidth]{{./results_puppets/1_output_u_init0_alphablur0_nu0.1_eta1_mu10_gamma0_eps0.05_time184.9985}.jpg}
&
\adjincludegraphics[valign=M,width=0.105\textwidth]{{./results_puppets/1_output_u_init0_alphablur0_nu0.1_eta1_mu10_gamma1_eps0.05_time182.1451}.jpg}
&
\adjincludegraphics[valign=M,width=0.105\textwidth]{{./results_puppets/1_output_u_init0_alphablur0_nu0.5_eta1_mu10_gamma0_eps0.05_time268.7268}.jpg}
&
\adjincludegraphics[valign=M,width=0.105\textwidth]{{./results_puppets/1_output_u_init0_alphablur0_nu0.5_eta1_mu10_gamma1_eps0.05_time232.8571}.jpg}
\\
\cmidrule{2-8}
& $100$ & 
\adjincludegraphics[valign=M,width=0.105\textwidth]{{./results_puppets/1_output_u_init0_alphablur0_nu0_eta1_mu100_gamma0_eps0.05_time3466.4818}.jpg}
&
\adjincludegraphics[valign=M,width=0.105\textwidth]{{./results_puppets/1_output_u_init0_alphablur0_nu0.5_eta1_mu10_gamma1_eps0.05_time232.8571}.jpg}
&
\adjincludegraphics[valign=M,width=0.105\textwidth]{{./results_puppets/1_output_u_init0_alphablur0_nu0.1_eta1_mu100_gamma0_eps0.05_time164.9425}.jpg}
&
\adjincludegraphics[valign=M,width=0.105\textwidth]{{./results_puppets/1_output_u_init0_alphablur0_nu0.1_eta1_mu100_gamma1_eps0.05_time154.0647}.jpg}
&
\adjincludegraphics[valign=M,width=0.105\textwidth]{{./results_puppets/1_output_u_init0_alphablur0_nu0.5_eta1_mu100_gamma0_eps0.05_time199.8359}.jpg}
&
\adjincludegraphics[valign=M,width=0.105\textwidth]{{./results_puppets/1_output_u_init0_alphablur0_nu0.1_eta1_mu100_gamma1_eps0.05_time154.0647}.jpg}
\\
\midrule
\multirow{2}{*}[-0.3cm]{
\adjincludegraphics[valign=M,width=0.105\textwidth]{{./results_puppets/1_composite}.jpg}}
& $10$ & 
\adjincludegraphics[valign=M,width=0.105\textwidth]{{./results_puppets/1_output_u_init1_alphablur0_nu0_eta1_mu10_gamma0_eps0.05_time546.4103}.jpg}
&
\adjincludegraphics[valign=M,width=0.105\textwidth]{{./results_puppets/1_output_u_init1_alphablur0_nu0_eta1_mu10_gamma1_eps0.05_time472.4963}.jpg}
&
\adjincludegraphics[valign=M,width=0.105\textwidth]{{./results_puppets/1_output_u_init1_alphablur0_nu0.1_eta1_mu10_gamma0_eps0.05_time88.0589}.jpg}
&
\adjincludegraphics[valign=M,width=0.105\textwidth]{{./results_puppets/1_output_u_init1_alphablur0_nu0.1_eta1_mu10_gamma1_eps0.05_time317.1639}.jpg}
&
\adjincludegraphics[valign=M,width=0.105\textwidth]{{./results_puppets/1_output_u_init1_alphablur0_nu0.5_eta1_mu10_gamma0_eps0.05_time201.6128}.jpg}
&
\adjincludegraphics[valign=M,width=0.105\textwidth]{{./results_puppets/1_output_u_init1_alphablur0_nu0.5_eta1_mu10_gamma1_eps0.05_time332.6901}.jpg}
\\
\cmidrule{2-8}
& $100$ & 
\adjincludegraphics[valign=M,width=0.105\textwidth]{{./results_puppets/1_output_u_init1_alphablur0_nu0_eta1_mu100_gamma0_eps0.05_time431.3187}.jpg}
&
\adjincludegraphics[valign=M,width=0.105\textwidth]{{./results_puppets/1_output_u_init1_alphablur0_nu0_eta1_mu100_gamma1_eps0.05_time461.8989}.jpg}
&
\adjincludegraphics[valign=M,width=0.105\textwidth]{{./results_puppets/1_output_u_init1_alphablur0_nu0.1_eta1_mu100_gamma0_eps0.05_time57.4665}.jpg}
&
\adjincludegraphics[valign=M,width=0.105\textwidth]{{./results_puppets/1_output_u_init1_alphablur0_nu0.1_eta1_mu100_gamma1_eps0.05_time146.8102}.jpg}
&
\adjincludegraphics[valign=M,width=0.105\textwidth]{{./results_puppets/1_output_u_init1_alphablur0_nu0.5_eta1_mu100_gamma0_eps0.05_time157.9888}.jpg}
&
\adjincludegraphics[valign=M,width=0.105\textwidth]{{./results_puppets/1_output_u_init1_alphablur0_nu0.5_eta1_mu100_gamma1_eps0.05_time185.294}.jpg}
\\
\midrule
\multirow{2}{*}[-0.3cm]{
\adjincludegraphics[valign=M,width=0.105\textwidth]{{./results_puppets/1_other_init}.jpg}
} 
& 10 &
\adjincludegraphics[valign=M,width=0.105\textwidth]{{./results_puppets/1_output_u_init2_alphablur0_nu0_eta1_mu10_gamma0_eps0.05_time668.9832}.jpg}
&
\adjincludegraphics[valign=M,width=0.105\textwidth]{{./results_puppets/1_output_u_init2_alphablur0_nu0_eta1_mu10_gamma1_eps0.05_time383.1681}.jpg}
&
\adjincludegraphics[valign=M,width=0.105\textwidth]{{./results_puppets/1_output_u_init2_alphablur0_nu0.1_eta1_mu10_gamma0_eps0.05_time195.6242}.jpg}
&
\adjincludegraphics[valign=M,width=0.105\textwidth]{{./results_puppets/1_output_u_init2_alphablur0_nu0.1_eta1_mu10_gamma1_eps0.05_time165.1121}.jpg}
&
\adjincludegraphics[valign=M,width=0.105\textwidth]{{./results_puppets/1_output_u_init2_alphablur0_nu0.5_eta1_mu10_gamma0_eps0.05_time240.6473}.jpg}
&
\adjincludegraphics[valign=M,width=0.105\textwidth]{{./results_puppets/1_output_u_init2_alphablur0_nu0.1_eta1_mu10_gamma1_eps0.05_time165.1121}.jpg}
\\
\cmidrule{2-8}
& 100 & 
\adjincludegraphics[valign=M,width=0.105\textwidth]{{./results_puppets/1_output_u_init2_alphablur0_nu0_eta1_mu100_gamma0_eps0.05_time2254.8568}.jpg}
&
\adjincludegraphics[valign=M,width=0.105\textwidth]{{./results_puppets/1_output_u_init2_alphablur0_nu0_eta1_mu100_gamma1_eps0.05_time517.3081}.jpg}
&
\adjincludegraphics[valign=M,width=0.105\textwidth]{{./results_puppets/1_output_u_init2_alphablur0_nu0.1_eta1_mu100_gamma0_eps0.05_time189.8453}.jpg}
&
\adjincludegraphics[valign=M,width=0.105\textwidth]{{./results_puppets/1_output_u_init2_alphablur0_nu0.1_eta1_mu100_gamma1_eps0.05_time151.5283}.jpg}
&
\adjincludegraphics[valign=M,width=0.105\textwidth]{{./results_puppets/1_output_u_init2_alphablur0_nu0.5_eta1_mu100_gamma0_eps0.05_time196.4417}.jpg}
&
\adjincludegraphics[valign=M,width=0.105\textwidth]{{./results_puppets/1_output_u_init2_alphablur0_nu0.5_eta1_mu100_gamma1_eps0.05_time177.2414}.jpg}
\\
\bottomrule
\end{tabularx}
\caption{Effect of different weighting parameters and initialization for \eqref{eq: joint model final} and images \ref{fig: puppets1}-\ref{fig: vangogh}.
}
\label{fig: puppets}
\end{figure*}

We consider the problem of image fusion of two images $f$ and $b$, \rev{respectively in Figure \ref{fig: dataset} (a)--(c)}, via a given $\alpha$-map.
In Figure \ref{fig: puppets} we show the results from our model for different parameters $\eta
\in\{0,0.1,0.5\}$, $\nu\in\{10,100\}$ and $\gamma
\in\{0,1\}$ and for a different starting image $u^0$, e.g.\ $u^0=f$, $u^0$ being the $\alpha$-convex combination of $f$ and $b$, or their average. \rev{Our tests suggest that a reasonable choice is $\mu=100\ \gg \eta=0.1,\gamma=1$.}
 Given the non-convexity of the joint model \eqref{eq: joint model final}, we notice that the initialisation plays a crucial role in terms of color preservation. 
Regarding the parameters, we choose the fidelity parameter $\mu$ to be very large with respect to $\eta$ and $\gamma$ in order to  to damp the effect of the regularisation and preserve structural information at every iteration well enough.
As expected, the combined effect of the $\alpha$-map in the fidelity term of $u$ and $v$ as well as the regularisation weight $\eta$ allows to control the level of the naturalness in the gluing process. 
Note, for instance, that for an increasing value of $\eta$, the staircasing artefact on $v$ due to the TV-type regularizer is more evident. 
Finally $\gamma$ controls the amount of the foreground information to be preserved, making its usage relevant for real applications where parts of the image need to remain intact.

\subsection{Face fusion}
\begin{figure*}
\centering
    \begin{subfigure}[t]{0.125\textwidth}\centering
    \captionsetup{justification=centering}
    \includegraphics[width=0.95\textwidth]{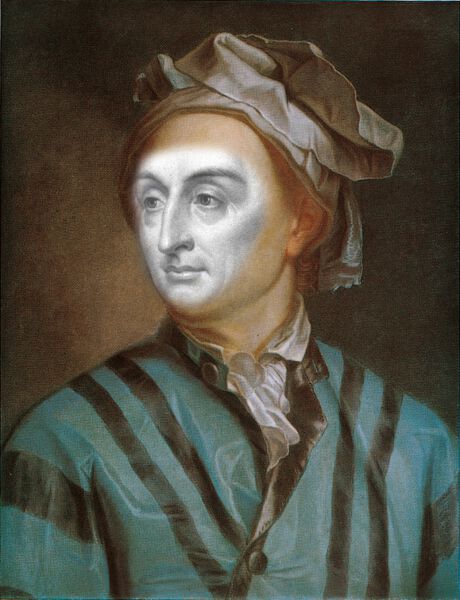}
    \caption{\rev{Direct cloning\\based on $\alpha$-map}}
    \end{subfigure}
    \,
    \begin{subfigure}[t]{0.125\textwidth}\centering
    \captionsetup{justification=centering}
    \includegraphics[width=0.95\textwidth]{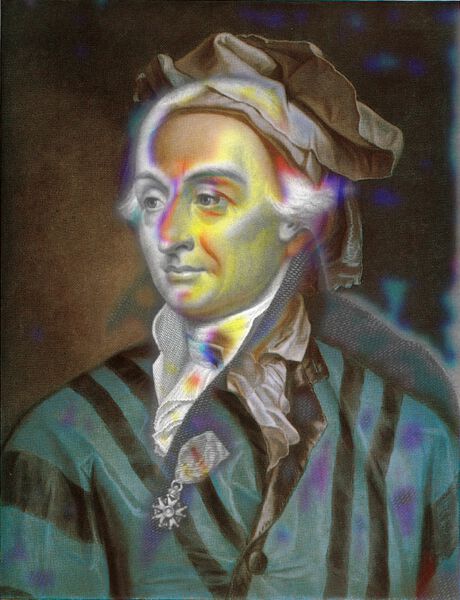}
    \caption{\rev{Structure-Aware\\\cite{SAIF} (0.85 s.)}}
    \end{subfigure}
    \,
    \begin{subfigure}[t]{0.125\textwidth}\centering
    \captionsetup{justification=centering}
    \includegraphics[width=0.95\textwidth]{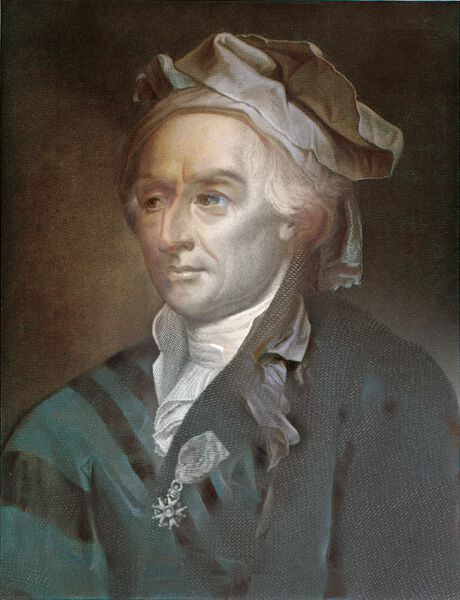}
    \caption{\rev{Guided\\\cite{GFF} (0.58 s.)}}
    \end{subfigure}
    \,
    \begin{subfigure}[t]{0.125\textwidth}\centering
    \captionsetup{justification=centering}
    \includegraphics[width=0.95\textwidth]{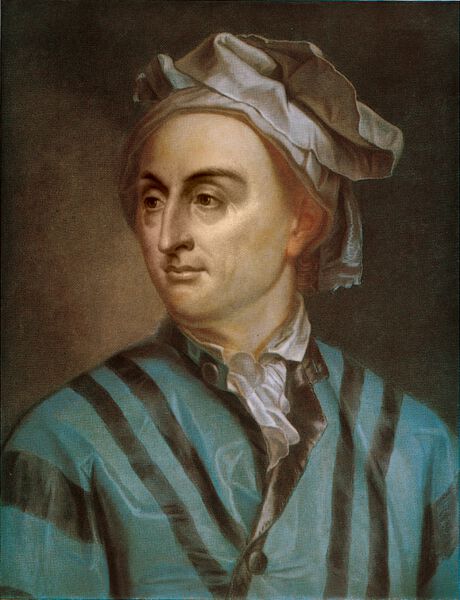}
    \caption{Seamless Poisson Editing \cite{Perez2003}
    \label{fig: face-fusion seamless}
    (34 s.)}
    \end{subfigure}
    \,
    \begin{subfigure}[t]{0.125\textwidth}\centering
    \captionsetup{justification=centering}
    \includegraphics[width=0.95\textwidth]{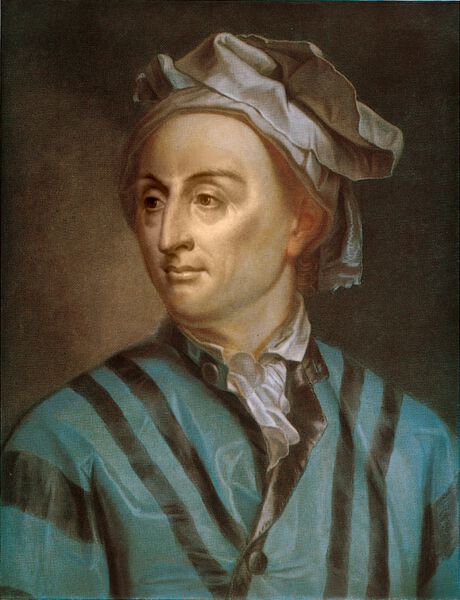}
    \caption{Non-linear  (w/out registr.) \cite{BenMolNosBurCreGilSch2017} (427 s.)\\ \rev{Eigenvectors:\\ $\small f\,(1^{\text{st}})+b\, (2^\text{nd}$-$12^{\text{th}})$.}}
     \label{fig: face-fusion nonlinear}
    \end{subfigure}
    \,
    \begin{subfigure}[t]{0.125\textwidth}\centering
    \captionsetup{justification=centering}
    \includegraphics[width=0.95\textwidth]{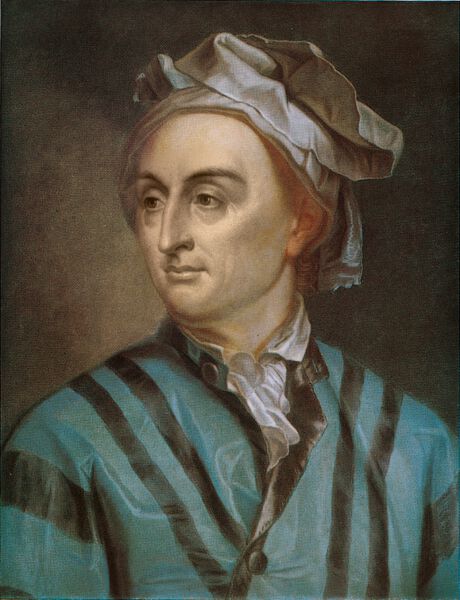}
    \caption{Osmosis \cite{Weickert2013}\\(79 s.), BiCGStab: \rev{\texttt{tol}=1\eu-05, \texttt{maxiter}=500}}
     \label{fig: face-fusion osmosis}
    \end{subfigure}
    \,
    \begin{subfigure}[t]{0.125\textwidth}\centering
    \captionsetup{justification=centering}
    \includegraphics[width=0.95\textwidth]{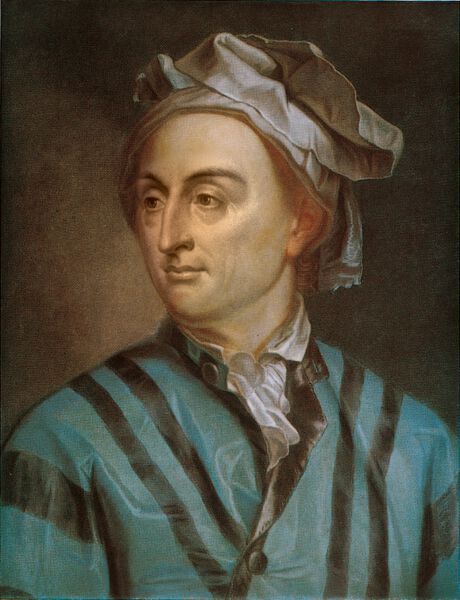}
    \caption{Ours \eqref{eq: joint model final}\\(469 s.), $u^0=f$\\ $(\eta,\mu,\gamma)=(0.01,10,1)$}
    \label{fig: face-fusion our approach}
    \end{subfigure}
    \caption{Comparison of different face fusion approaches \rev{and cputime} (better viewed in color).}
    \label{fig: face fusion}
    \end{figure*}

\begin{figure*}
\centering
 \begin{subfigure}[t]{0.125\textwidth}\centering
    \captionsetup{justification=centering}
    \includegraphics[width=0.9\textwidth,trim=4cm 9.5cm 6cm 3.5cm,clip=true]{{./facefusion/4_direct_cloning}.jpg}
    \caption{\rev{Direct matting (zoom)}}
    \end{subfigure}
    \,
     \begin{subfigure}[t]{0.125\textwidth}\centering
    \captionsetup{justification=centering}
    \includegraphics[width=0.9\textwidth,trim=4cm 9.5cm 6cm 3.5cm,clip=true]{{./facefusion/4_SAIF_time0.85}.jpg}
    \caption{\rev{\cite{SAIF} (zoom)}}
    \end{subfigure}
    \,
     \begin{subfigure}[t]{0.125\textwidth}\centering
    \captionsetup{justification=centering}
    \includegraphics[width=0.9\textwidth,trim=4cm 9.5cm 6cm 3.5cm,clip=true]{{./facefusion/4_GFF_time0.58}.jpg}
    \caption{\rev{\cite{GFF} (zoom)}}
    \end{subfigure}
    \,
    \begin{subfigure}[t]{0.125\textwidth}\centering
    \captionsetup{justification=centering}
    \includegraphics[width=0.9\textwidth,trim=4cm 9.5cm 6cm 3.5cm,clip=true]{{./facefusion/4_poisson_editing_time34.14}.jpg}
    \caption{Poisson (zoom)}
    \end{subfigure}
    \,
    \begin{subfigure}[t]{0.125\textwidth}\centering
    \captionsetup{justification=centering}
    \includegraphics[width=0.9\textwidth,trim=4cm 9.5cm 6cm 3.5cm,clip=true]{{./facefusion/4_k1_p12_nonlinear_time427.44}.jpg}
    \caption{Non-linear (zoom)}
    \end{subfigure}
    \,
    \begin{subfigure}[t]{0.125\textwidth}\centering
    \captionsetup{justification=centering}
    \includegraphics[width=0.9\textwidth,trim=4cm 9.5cm 6cm 3.5cm,clip=true]{{./facefusion/4_osmosis_fusion_T10000_dt1000_tol1e-05_maxit500_theta1_time79.67}.jpg}
    \caption{Osmosis (zoom)}
    \end{subfigure}
    \,
    \begin{subfigure}[t]{0.125\textwidth}\centering
    \captionsetup{justification=centering}
    \includegraphics[width=0.9\textwidth,trim=4cm 9.5cm 6cm 3.5cm,clip=true]{{./facefusion/4_output_u_init0_alphablur5_nu0.01_eta1_mu10_gamma1_eps0.05_time469.8475}.jpg}
    \caption{Ours \eqref{eq: joint model final} (zoom)}
    \end{subfigure}
    \\
    \begin{subfigure}[t]{0.125\textwidth}\centering
    \captionsetup{justification=centering}
    \includegraphics[width=0.9\textwidth,trim=4cm 9.5cm 6cm 3.5cm,clip=true]{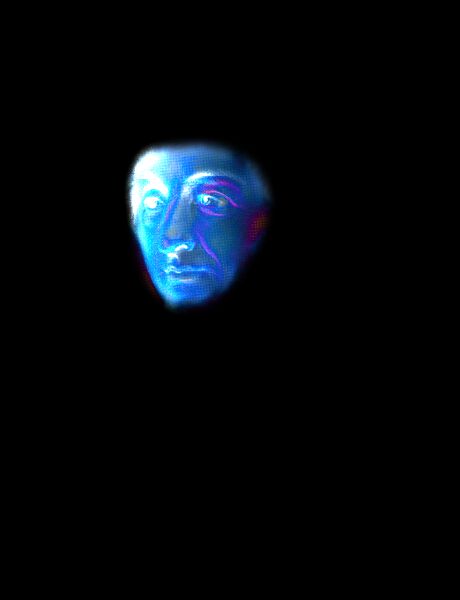}
    \caption{\rev{Error \eqref{eq: chroma error}\\ (Direct vs.\ $f$)
    }}
    \end{subfigure}
    \,
    \begin{subfigure}[t]{0.125\textwidth}\centering
    \captionsetup{justification=centering}
    \includegraphics[width=0.9\textwidth,trim=4cm 9.5cm 6cm 3.5cm,clip=true]{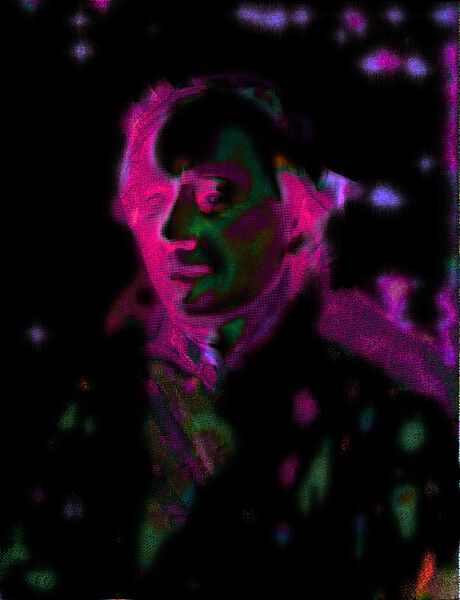}
    \caption{\rev{Error \eqref{eq: chroma error}\\ (\cite{SAIF} vs.\ $f$)}}
    \end{subfigure}
    \,
    \begin{subfigure}[t]{0.125\textwidth}\centering
    \captionsetup{justification=centering}
    \includegraphics[width=0.9\textwidth,trim=4cm 9.5cm 6cm 3.5cm,clip=true]{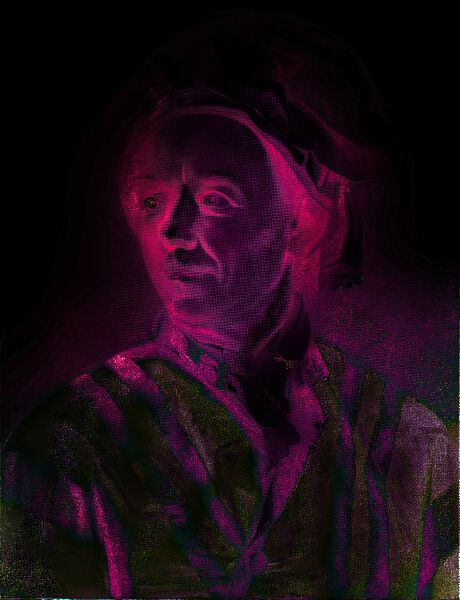}
    \caption{\rev{Error \eqref{eq: chroma error}\\ (\cite{GFF} vs.\ $f$)}}
    \end{subfigure}
    \,
    \begin{subfigure}[t]{0.125\textwidth}\centering
    \captionsetup{justification=centering}
    \includegraphics[width=0.9\textwidth,trim=4cm 9.5cm 6cm 3.5cm,clip=true]{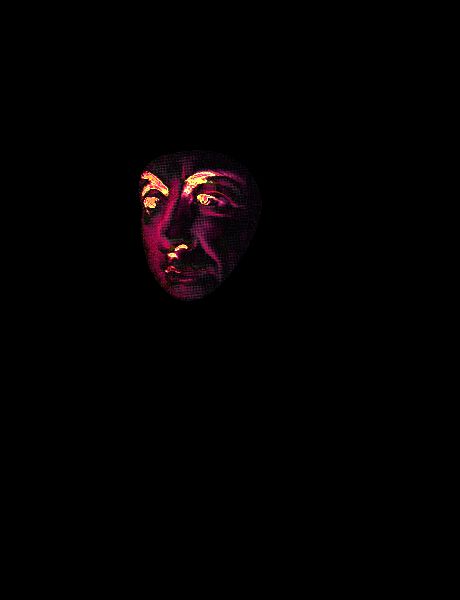}
    \caption{Error \eqref{eq: chroma error} (Poisson vs.\ $f$)}
    \end{subfigure}
    \,    
    \begin{subfigure}[t]{0.125\textwidth}\centering
    \captionsetup{justification=centering}
    \includegraphics[width=0.9\textwidth,trim=4cm 9.5cm 6cm 3.5cm,clip=true]{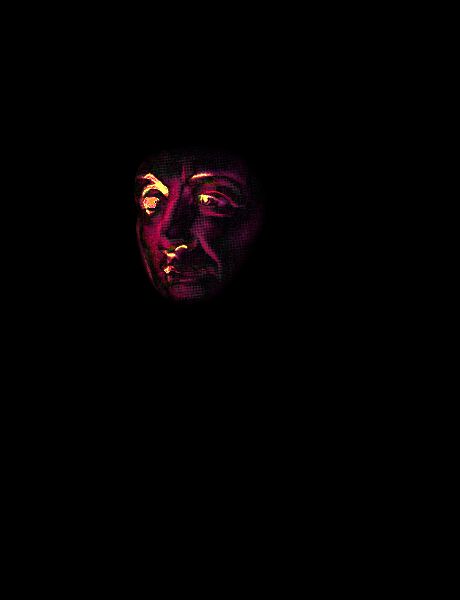}
    \caption{Error \eqref{eq: chroma error} (Non-linear vs.\ $f$)}
    \end{subfigure}
    \,
    \begin{subfigure}[t]{0.125\textwidth}\centering
    \captionsetup{justification=centering}
    \includegraphics[width=0.9\textwidth,trim=4cm 9.5cm 6cm 3.5cm,clip=true]{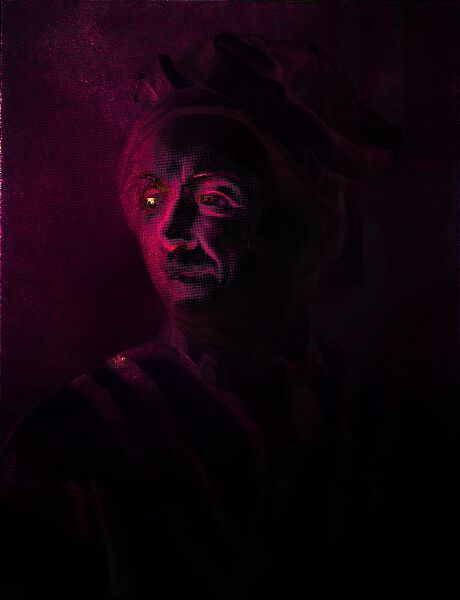}
    \caption{Error \eqref{eq: chroma error}  (Osmosis vs.\ $f$)}
    \end{subfigure}
    \,        
    \begin{subfigure}[t]{0.125\textwidth}\centering
    \captionsetup{justification=centering}
    \includegraphics[width=0.9\textwidth,trim=4cm 9.5cm 6cm 3.5cm,clip=true]{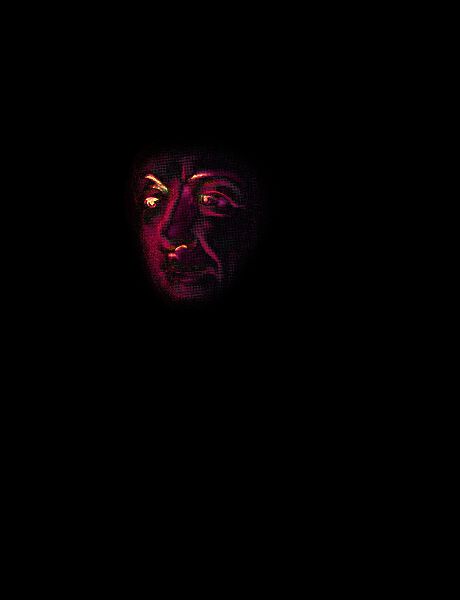}
    \caption{Error \eqref{eq: chroma error}\\(Ours vs.\ $f$)}
    \end{subfigure}
    \caption{Zoom of face fusion results from Figure \ref{fig: face fusion} and chromaticity error \eqref{eq: chroma error}.}
    \label{fig: zoom face-fusion}
\end{figure*}

We now focus on a face fusion application.
We compare our model\rev{, for which the $\alpha$-map has been obtained by convolution with a Gaussian kernel of standard deviation $\sigma=5$,} with other approaches, \rev{using the parameters suggested by the authors in their original papers}: the naive direct \rev{matting, the structure-aware fusion \cite{SAIF}, the guided fusion \cite{GFF}}, the seamless Poisson editing \cite{Perez2003}, the non-linear fusion \cite{BenMolNosBurCreGilSch2017} (without pre-processing registration), the classic osmosis filter \cite{Weickert2013} (with $\dbold$ being the average of the drift associated to $f$ and to $b$ on the transition zone).
In order to compare the different performances, we introduce a \emph{chromaticity error} defined in terms of the geometric chromaticity introduced in \cite{FinHorDre2006} to highlight both the changes in the image structure and in the color information between the result and the starting image.
For a \rev{general} RGB image $z$ we define the Geometric Chromaticity Mean of $z$ as $\text{GCM}(z):=\sqrt[3]{z_R z_G z_B}$,  \rev{so that the pixel-wise chromaticity error between two given images $u^1$ and $u^2$ is defined as}
\begingroup\makeatletter\def\f@size{9.5}\check@mathfonts
\def\maketag@@@#1{\hbox{\m@th\normalsize\normalfont#1}}
\rev{\begin{equation}
\text{err}(u^1,u^2) = \abs{\frac{u_{R,G,B}^1}{\text{GCM}(u^1)}-\frac{u^2_{R,G,B}}{\text{GCM}(u^2)}}.
\label{eq: chroma error}    
\end{equation}
}
\endgroup
We report in Fig.\ \ref{fig: face fusion} (zoom in Fig.\ \ref{fig: zoom face-fusion}) the different approaches  used to fuse images in Fig.\ \ref{fig: dataset} (d) and (e). 
The result obtained with our approach is reported in Fig.\ \ref{fig: face-fusion our approach}: it clearly produces a better skin tone than the seamless Poisson editing model, see Fig.\ \ref{fig: face-fusion seamless}, and than the pure osmosis model, see Fig.\ \ref{fig: face-fusion osmosis}. 
The result obtained by the non-linear fusion model in Fig.\ \ref{fig: face-fusion nonlinear} is comparable with ours, but from a computational point of view it requires to manually select the eigenvectors to fuse, resulting either into a sharpening effect or in undesired results. 
Furthermore, we note that the chromaticity error map \eqref{eq: chroma error} displayed in Fig.\ \ref{fig: zoom face-fusion} highlights that the biggest errors are generally localized where image structures change: our approach mitigates such errors due to the control of the fusion via the blurred $\alpha$-map and the regularization on $v$, see in particular the eyes and the cheeks.
\rev{Quantitatively, none of the existing no-reference image quality score tested (
\revbis{BRISQUE \cite{brisque}, FMI \cite{fmi}}) 
in Table \ref{tab: quantitative} (average on color channels) were able to highlight significant differences in the results of Fig.\ \ref{fig: face fusion} as $\text{err}(\cdot,\cdot)$ 
\revbis{in} \eqref{eq: chroma error} (note 
that good values of 
\revbis{BRISQUE and FMI} do not correspond to a visually pleasant results).}

\begin{table}[!htb]\scriptsize
\scriptsize
\begin{tabularx}{0.485\textwidth}{X|c|c|c}
\toprule\scriptsize
 & $\norm{\text{err}(u,f)}_2$ & BRISQUE \cite{brisque} & FMI \cite{fmi}\\
\midrule
Direct matting               & 0.0178 & 30.3 & 0.895\\
Structure-Aware \cite{SAIF}  & 0.0659 & 30.9 & 0.879\\
Guided \cite{GFF}            & 0.0790 & 32.6 & 0.883\\
Poisson \cite{Perez2003}     & 0.0111 & 30.2 & \textbf{0.896}\\
Spectral \cite{BenMolNosBurCreGilSch2017} & 0.0087 & \textbf{30.1} & 0.896\\
Osmosis \cite{Weickert2013} & 0.0452 & 30.2 & 0.895\\
Proposed & \textbf{0.0062} & \textbf{30.1} & 0.895\\
\bottomrule
\end{tabularx}
\caption{\revbis{Evaluation of image fusion methods for face fusion in Figure \ref{fig: face fusion} (average on color channels, best results in bold).}}
\label{tab: quantitative}
\end{table}

\subsection{Multi-modal fusion in Cultural Heritage}
The image in Figure \ref{fig: fitzwilliam VIS} comes from a deluxe copy
made in 1290 of \emph{La Somme le roi} by Master Honor\'e's. This is a "survey for the king" on moral conduct, that  consists of five treatises, probably made for Philip IV of France (1285-1314) and his queen, Jeanne of Navarre \cite{RouRou2000}.
The leaf has size $180\times 125$mm and the miniature $135\times90$mm. 

\begin{figure*}
    \centering
    \begin{subfigure}[t]{0.265\textwidth}\centering
    \captionsetup{justification=centering}
    \includegraphics[width=0.835\textwidth]{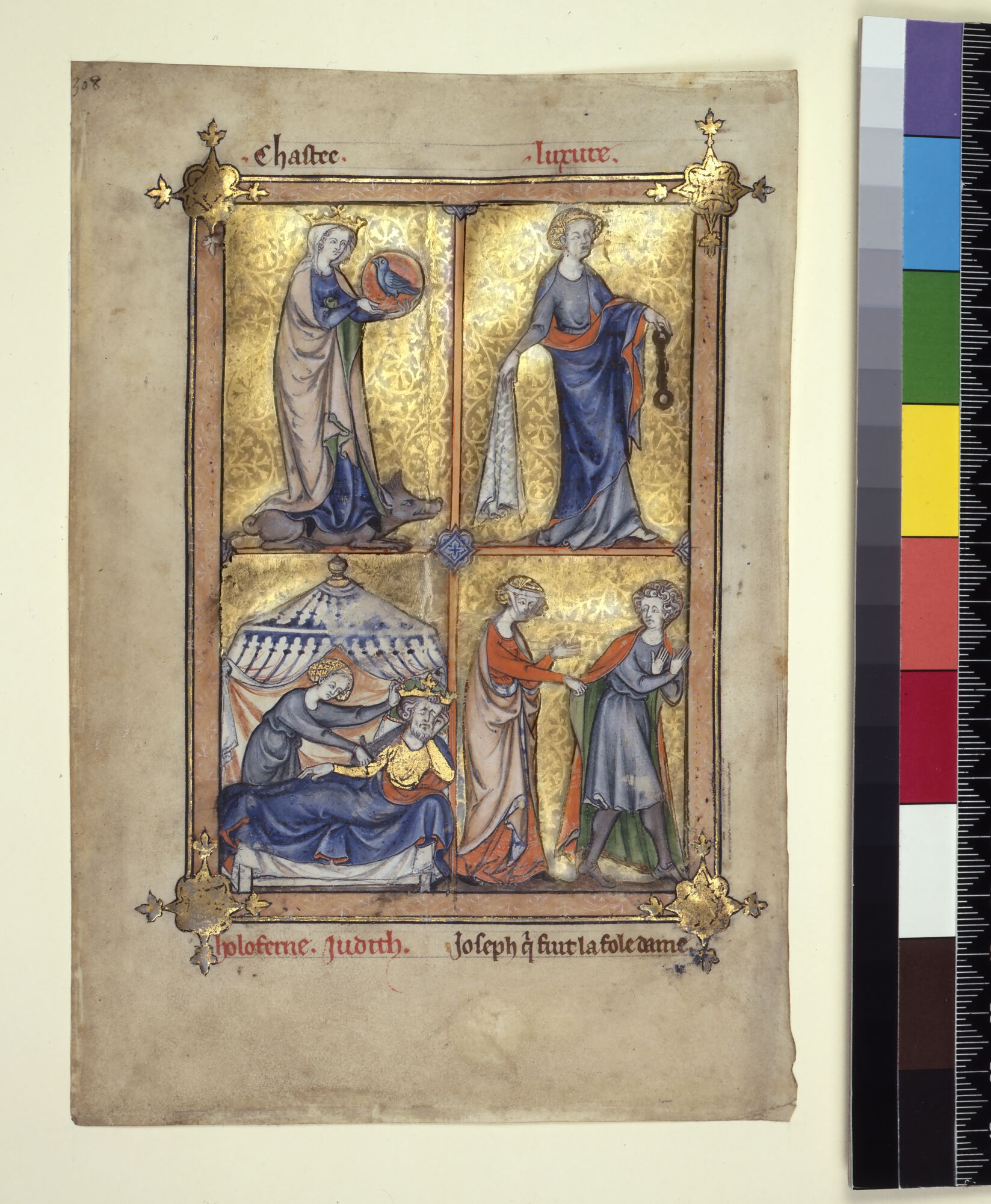}
    \caption{
    Ours \eqref{eq: joint model final} (3921 s.),\\$(\eta,\mu,\gamma)=(0.1,10,0.01)$}
    \label{fig: resultCH1}
    \end{subfigure}
    \,
    \begin{subfigure}[t]{0.225\textwidth}\centering
    \captionsetup{justification=centering}
    \includegraphics[width=0.835\textwidth,trim=25cm 38cm 18cm 11cm,clip=true]{{./manuscripts_fitz/3_foreground}.jpg}
    \caption{Zoom of $f$}
    \end{subfigure}
    \,
    \begin{subfigure}[t]{0.225\textwidth}\centering
    \captionsetup{justification=centering}
    \includegraphics[width=0.835\textwidth,trim=25cm 38cm 18cm 11cm,clip=true]{{./manuscripts_fitz/3_background}.jpg}
    \caption{Zoom of $b$}
    \end{subfigure}
    \,
    \begin{subfigure}[t]{0.225\textwidth}\centering
    \captionsetup{justification=centering}
    \includegraphics[width=0.835\textwidth,trim=25cm 38cm 18cm 11cm,clip=true]{{./manuscripts_fitz/3_output_u_init0_alphablur1_nu0.1_eta1_mu10_gamma0.01_eps0.05_time10861.0258}.jpg}
    \caption{
    Zoom of our result in \ref{fig: resultCH1}}
    \end{subfigure}
    \caption{Application to Cultural Heritage Images. Initialisation $u^0=f$.}
    \label{fig: fusion fitzwilliam}
\end{figure*}

The infrared diagnostic at \emph{The Fitzwilliam Museum} (Cambridge, UK) highlighted the presence of a background texture in the gold leaf, damaged throughout centuries, see Figure \ref{fig: fitzwilliam IR}. 
In view of fusing the multi-modal images so as to unveil the lost background, we \rev{aligned the two images using} SIFT \cite{SIFT} and computed the $\alpha$-map via \cite{Chen2012}.
Then, we apply our model \eqref{eq: joint model final} obtaining the result in Figure \ref{fig: fusion fitzwilliam}, with zoomed details. 
In the fusion result, the golden background is well-reconstructed without significantly touching the foreground information.

\begin{figure}
    \centering
    \begin{subfigure}[t]{0.155\textwidth}\centering
    \captionsetup{justification=centering}
    \includegraphics[width=0.925\textwidth,trim=0cm 1.95cm 0cm 1.95cm,clip=true]{{./results_vr/2_foreground_low}.jpg}
    \caption{Foreground}
    \end{subfigure}
    \begin{subfigure}[t]{0.155\textwidth}\centering
    \captionsetup{justification=centering}
    \includegraphics[width=0.925\textwidth,trim=0cm 1.5cm 0cm 1.5cm,clip=true]{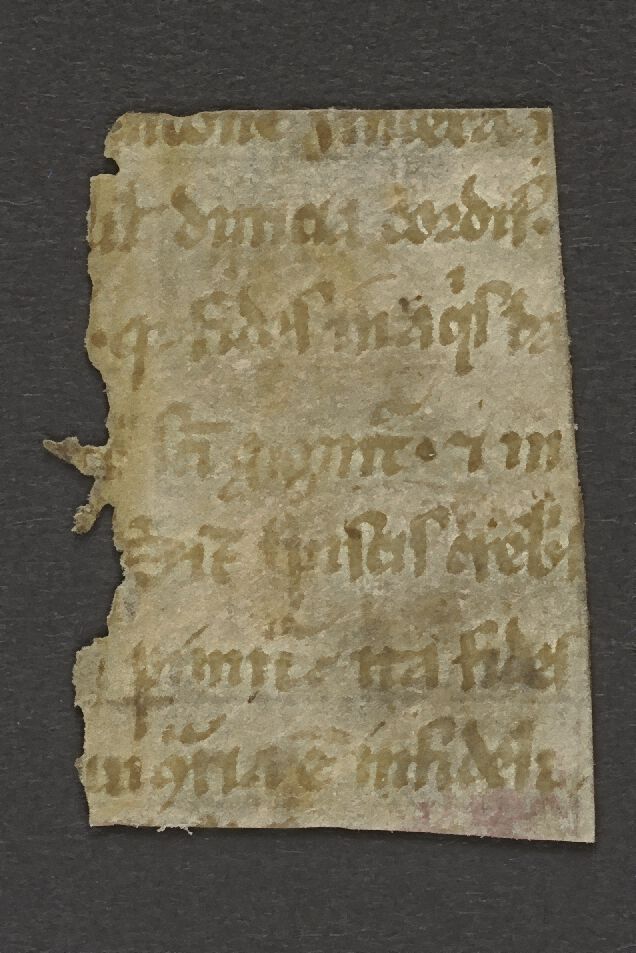}
    \caption{$(\eta,\mu,\gamma)=(0,10,0.1)$}
    \label{fig: text2}
    \end{subfigure}
    \begin{subfigure}[t]{0.155\textwidth}\centering
    \captionsetup{justification=centering}
    \includegraphics[width=0.925\textwidth,trim=0cm 1.5cm 0cm 1.5cm,clip=true]{{./results_vr/2_output_u_init0_alphablur0_nu0.1_eta1_mu10_gamma0.1_eps0.05_time330.9714}.jpg}
    \caption{$(\eta,\mu,\gamma)=(0.1,10,0.1)$}
    \label{fig: text3}
    \end{subfigure}
    \\
    \begin{subfigure}[t]{0.155\textwidth}\centering
    \captionsetup{justification=centering}
    \includegraphics[width=0.925\textwidth,trim=13cm 6cm 2cm 31.2cm,clip=true]{{./results_vr/2_foreground_low}.jpg}
    \caption{Foreground (zoom)}
    \end{subfigure}
    \begin{subfigure}[t]{0.155\textwidth}\centering
    \captionsetup{justification=centering}
    \includegraphics[width=0.925\textwidth,trim=11cm 4.8cm 1.5cm 25cm,clip=true]{{./results_vr/2_output_u_init0_alphablur0_nu0_eta1_mu10_gamma0.1_eps0.05_time384.7182}.jpg}
    \caption{Zoom of \ref{fig: text2}}
    \end{subfigure}
    \begin{subfigure}[t]{0.155\textwidth}\centering
    \captionsetup{justification=centering}
    \includegraphics[width=0.925\textwidth,trim=11cm 4.8cm 1.5cm 25cm,clip=true]{{./results_vr/2_output_u_init0_alphablur0_nu0.1_eta1_mu10_gamma0.1_eps0.05_time330.9714}.jpg}
    \caption{Zoom of \ref{fig: text3}}
    \end{subfigure}
    \caption{Text enhancing from a manuscript.
    }
    \label{fig: Verona manuscript}
\end{figure}

Figure \ref{fig: Verona VIS} shows a manuscript from \emph{Biblioteca Capitolare} (Verona, Italy), where the text is highly damaged but still visible after infrared inspection (Figure \ref{fig: Verona IR}). 
We can similarly apply our model \eqref{eq: joint model final} with the purpose of text enhancing by means of fusing the structural infrared information with the visible colors of the manuscript. 
Since multi-modal infrared acquisition can come with noise, we test different regularising parameter $\eta$ for an heuristically fixed choice of $\mu$ and $\gamma$.
In Figure \ref{fig: Verona manuscript}, we report different results showing that the larger $\eta$ is, the more the typical staircasing effect is visible in the fusion of the structural information. This effect, is in fact desirable in this application for enhancing the edges of the letters. 

We remark that despite the availability of \rev{several metrics assessing the quality of image fusion such as the ones used for comparison in Table \ref{tab: quantitative}, for} Cultural Heritage applications the validation of the optimal result is often performed by expert users, who are able to take into account not only image features, but possible other characteristics, \rev{e.g.\ author style, historic period and illumination}.

\section{Conclusions}
In this work we introduced a new model for non-linear image fusion. The fusion process is driven by an \emph{osmosis} term and involves also a regularization term and a fidelity term on the data. 
After motivating the model from a theoretical point of view, we discussed its numerical implementation and we applied it to solve different problems in the context of data fusion.
In particular, we highlighted the potentialities of the model for applications in Cultural Heritage imaging, 
from the unveiling of missed structures to text enhancing.

As future directions, we envisage the use of structural information not only in the drift field but also to build an anisotropy metric in the model, e.g.\ via the directional osmosis energy studied in \cite{Parisotto_2019}.
The fusion could also be performed in the spectral domain \rev{and faster solvers could be investigated so as to speed up the resolution of the optimization problem, such as, e.g.\ stochastic batch methods for non-convex composite optimization \cite{Ghadimi2016}}. 
At the same time, it could be of interest to force the regularizer to act only onto one of the two images to be fused, i.e.\ assuming that one image comes with noise and the other is noise-free. 
\revbis{Future work shall also be dedicated to extending the definition of the chromaticity error \eqref{eq: chroma error} to the non-reference dependent case, similarly to BRISQUE \cite{brisque} and FMI \cite{fmi}. Moreover, a quantitative assessment of the smoothness at the boundary of the fused images is also envisaged to promote a natural reconstruction. 
Finally, the learning of the osmosis operator directly from the data with modern deep learning approaches is envisaged so as to fuse} many more image features, see, e.g., \cite{Liu2018} for a review.

\ifCLASSOPTIONcompsoc
  \section*{Acknowledgements}
\else
  \section*{Acknowledgements}
\fi
The authors thank Prof.\ Joachim Weickert for fruitful discussions related to this work.
SP acknowledges the Leverhulme Trust project on ``Unveiling the invisible: mathematics for conservation in arts and humanities''. 
LC acknowledges the support given by the Fondation Math\'ematique Jacques Hadamard (FMJH).
 The CH diagnostics were supported by Dr Stella Panayotova and Dr Paola Ricciardi for Fig.\ \ref{fig: fitzwilliam IR} (Fitzwilliam Museum, Univ.\ of Cambridge), Dr Claudia Daffara (Univ.\ of Verona) for Fig.\ \ref{fig: Verona IR}. CBS acknowledges support from the Leverhulme Trust project on Breaking the non-convexity barrier, the Philip Leverhulme Prize, the EPSRC grant No. EP/M00483X/1, the EPSRC Centre No. EP/N014588/1, the European Union Horizon 2020 research and innovation programmes under the Marie Sk\l{}odowska-Curie grant agreement  No. 691070 CHiPS, the Cantab Capital Institute for the Mathematics of Information and the Alan Turing Institute.
This project is also funded by the EU Horizon 2020 research and innovation programme NoMADS (Marie Skłodowska-Curie grant agreement No 777826) and with financial support from the French Research Agency through the PostProdLEAP project (ANR-19-CE23-0027-01).

\vspace{-0.5em}
\appendix  \label{appendix}

\rev{
We detail here the derivation of \eqref{eq: grad v g2}:
\begin{footnotesize}
\[
\begin{aligned}
\pderiv{}{v}\left(\int_\Omega v \abs{\grad \left(\frac{u}{v}\right)}^2\diff \xbold
\right)
    &=
    \text{(a)}+\text{(b)}+\text{(c)},\quad\text{ where:}
\end{aligned}
\]
\end{footnotesize}
}
\begin{scriptsize}
\[
\begin{aligned}
\text{(a)}
&=
\pderiv{}{v}
\int_\Omega \frac{\grad u \cdot \grad u}{v} 
\diff\xbold
= 
\int_\Omega -\frac{\grad u \cdot \grad u}{v^2} 
\diff\xbold;
\\
(\text{b}) 
&= 
\pderiv{}{v}
\int_\Omega \frac{2u}{v^2}  \grad u\cdot \grad v 
\diff\xbold\\
&= 
\int_\Omega 
\pderiv{}{v} \left(\frac{2u}{v^2}\right) \grad u \cdot \grad v + \frac{2u}{v^2}\pderiv{}{v}\left(\grad u\cdot\grad v\right)
\diff\xbold\\
&=
\int_\Omega 
-\frac{4u}{v^3}\ \grad u \cdot \grad v  -\div\left(\frac{2u}{v^2}\grad u\right)
\diff\xbold;
\\
(\text{c}) 
&= 
\pderiv{}{v}\int_\Omega \frac{u^2}{v^3}  \grad v\cdot \grad v \diff\xbold \\
&=\int_\Omega \pderiv{}{v} \left(\frac{u^2}{v^3}\right) \grad v\cdot \grad v  + \frac{u^2}{v^3}\pderiv{}{v}\left(\grad v\cdot\grad v\right)\diff\xbold\\
&=
\int_\Omega 
-3\frac{u^2}{v^4}\grad v\cdot\grad v - v\div\left(\frac{u^2}{v^3}\grad v \right)\diff\xbold.
\end{aligned}
\]
\vspace{-2em}
\end{scriptsize}

\bibliographystyle{IEEEtranS}
\bibliography{biblio.bib}

\begin{thebibliography}{10}
\providecommand{\url}[1]{#1}
\csname url@samestyle\endcsname
\providecommand{\newblock}{\relax}
\providecommand{\bibinfo}[2]{#2}
\providecommand{\BIBentrySTDinterwordspacing}{\spaceskip=0pt\relax}
\providecommand{\BIBentryALTinterwordstretchfactor}{4}
\providecommand{\BIBentryALTinterwordspacing}{\spaceskip=\fontdimen2\font plus
\BIBentryALTinterwordstretchfactor\fontdimen3\font minus
  \fontdimen4\font\relax}
\providecommand{\BIBforeignlanguage}[2]{{%
\expandafter\ifx\csname l@#1\endcsname\relax
\typeout{** WARNING: IEEEtranS.bst: No hyphenation pattern has been}%
\typeout{** loaded for the language `#1'. Using the pattern for}%
\typeout{** the default language instead.}%
\else
\language=\csname l@#1\endcsname
\fi
#2}}
\providecommand{\BIBdecl}{\relax}
\BIBdecl

\bibitem{Amolins2007}
K.~Amolins, Y.~Zhang, and P.~Dare, ``Wavelet based image fusion techniques - an
  introduction, review and comparison,'' \emph{ISPRS J.\ Photogramm.\ Remote
  Sens.}, vol. 62(4), 2007.

\bibitem{Ancuti2016night}
C.~Ancuti, C.~O. Ancuti, C.~De~Vleeschouwer, and A.~C. Bovik, ``Night-time
  dehazing by fusion,'' in \emph{IEEE ICIP}, 2016, pp. 2256--2260.

\bibitem{Ancuti2012enhancing}
C.~Ancuti, C.~O. Ancuti, T.~Haber, and P.~Bekaert, ``Enhancing underwater
  images and videos by fusion,'' in \emph{IEEE CVPR}, 2012, pp. 81--88.

\bibitem{BenMolNosBurCreGilSch2017}
M.~Benning, M.~M{\"o}ller, R.~Z. Nossek, M.~Burger, D.~Cremers, G.~Gilboa, and
  C.-B. Sch{\"o}nlieb, ``Nonlinear spectral image fusion,'' in \emph{SSVM},
  2017, pp. 41--53.

\bibitem{Burt83}
P.~J. Burt and E.~H. Adelson, ``A multiresolution spline with application to
  image mosaics,'' \emph{ACM Trans.\ Graph.}, vol. 2(4), pp. 217--236, 1983.

\bibitem{Calatroni2017}
L.~Calatroni, C.~Estatico, N.~Garibaldi, and S.~Parisotto, ``Alternating
  {D}irection {I}mplicit ({ADI}) schemes for a {PDE}-based image osmosis
  model,'' \emph{J.\ Phys.\ Conf.\ Ser.}, vol. 904(1), p. 012014, 2017.

\bibitem{Calatroni2018}
L.~Calatroni, M.~d'Autume, R.~Hocking, S.~Panayotova, S.~Parisotto,
  P.~Ricciardi, and C.-B. Sch{\"o}nlieb, ``Unveiling the invisible:
  mathematical methods for restoring and interpreting illuminated
  manuscripts,'' \emph{Heritage Science}, vol. 6(1), p.~56, 2018.

\bibitem{ChaPoc2011}
A.~Chambolle and T.~Pock, ``A first-order primal-dual algorithm for convex
  problems with applications to imaging,'' \emph{J.\ Math.\ Imaging Vis.}, vol.
  40(1), pp. 120--145, 2011.

\bibitem{Chen2012}
Q.~Chen, D.~Li, and C.-K. Tang, ``Knn matting,'' \emph{IEEE Trans.\ Pattern
  Anal.\ Mach.\ Intell.}, vol. 35(9), pp. 2175--2188, 2013.

\bibitem{Saad2013}
S.~Darwish, ``Multi-level fuzzy contourlet-based image fusion for medical
  applications,'' \emph{Image Processing, IET}, vol.~7, pp. 694--700, 10 2013.

\bibitem{AutMorLlo18}
M.~d'Autume, J.~Morel, and E.~Meinhardt{-}Llopis, ``A flexible solution to the
  osmosis equation for seamless cloning and shadow removal,'' in \emph{IEEE
  ICIP}, 2018, pp. 2147--2151.

\bibitem{Fattal2002}
R.~Fattal, D.~Lischinski, and M.~Werman, ``Gradient domain high dynamic range
  compression,'' in \emph{ACM Trans.\ Graph.}, vol. 21(3), 2002.

\bibitem{Mirebeau2014}
J.~Fehrenbach and J.-M. Mirebeau, ``Sparse non-negative stencils for
  anisotropic diffusion,'' \emph{J. Math. Imaging Vis.}, vol.~49, pp. 123--147,
  2014.

\bibitem{Feng2019}
R.~Feng, Q.~Du, X.~Li, and H.~Shen, ``Robust registration for remote sensing
  images by combining and localizing feature- and area-based methods,''
  \emph{ISPRS J.\ Photogramm.\ Remote Sens.}, vol. 151, 2019.

\bibitem{FinHorDre2006}
G.~D. {Finlayson}, S.~D. {Hordley}, {Cheng Lu}, and M.~S. {Drew}, ``On the
  removal of shadows from images,'' \emph{IEEE Trans.\ Pattern Anal.\ Mach.\
  Intell.}, vol. 28(1), pp. 59--68, 2006.

\bibitem{Gabarda2007}
S.~Gabarda and G.~Cristóbal, ``Cloud covering denoising through image
  fusion,'' \emph{Image Vision Comput.}, vol. 25(5), pp. 523 -- 530, 2007.

\bibitem{Ghadimi2016}
S.~Ghadimi, G.~Lan, and H.~Zhang, ``Mini-batch stochastic approximation methods
  for nonconvex stochastic composite optimization,'' \emph{Mathematical
  Programming}, vol. 155, no.~1, pp. 267--305, 2016.

\bibitem{HafnerPHD}
D.~Hafner, ``Variational image fusion,'' Ph.D. dissertation, Faculties of
  Mathematics and Computer Science, Saarland University, 2017.

\bibitem{Hagenburg2012}
K.~Hagenburg, M.~Breu{\ss}, J.~Weickert, and O.~Vogel, ``Novel schemes for
  hyperbolic pdes using osmosis filters from visual computing,'' in
  \emph{SSVM}, 2011, pp. 532--543.

\bibitem{fmi}
M.~{Haghighat} and M.~A. {Razian}, ``Fast-fmi: Non-reference image fusion
  metric,'' in \emph{IEEE AICT}, 2014, pp. 1--3.

\bibitem{Hait2019}
E.~{Hait} and G.~{Gilboa}, ``Spectral total-variation local scale signatures
  for image manipulation and fusion,'' \emph{IEEE Trans. on Image Processing},
  vol. 28(2), pp. 880--895, 2019.

\bibitem{James2014medical}
A.~P. James and B.~V. Dasarathy, ``Medical image fusion: A survey of the state
  of the art,'' \emph{Inform.\ Fusion}, vol.~19, pp. 4--19, 2014.

\bibitem{Kim2011}
Y.~{Kim}, C.~{Lee}, D.~{Han}, Y.~{Kim}, and Y.~{Kim}, ``Improved
  additive-wavelet image fusion,'' \emph{IEEE Geosci.\ Remote Sens.\ Lett.},
  vol. 8(2), 2011.

\bibitem{Li2016}
F.~Li and T.~Zeng, ``Variational image fusion with first and second-order
  gradient information,'' \emph{J.\ Comp.\ Math.}, vol.~34, pp. 200--222, 03
  2016.

\bibitem{GFF}
S.~{Li}, X.~{Kang}, and J.~{Hu}, ``Image fusion with guided filtering,''
  \emph{IEEE Trans.\ Image Process.}, vol. 22(7), pp. 2864--2875, 2013.

\bibitem{Li2017}
S.~Li, X.~Kang, L.~Fang, J.~Hu, and H.~Yin, ``Pixel-level image fusion: A
  survey of the state of the art,'' \emph{Inform.\ Fusion}, vol.~33, 2017.

\bibitem{Li2008multifocus}
S.~Li and B.~Yang, ``Multifocus image fusion by combining curvelet and wavelet
  transform,'' \emph{Pat. Recogn. Lett.}, vol. 29(9), pp. 1295--1301, 2008.

\bibitem{SAIF}
W.~Li, Y.~Xie, H.~Zhou, Y.~Han, and K.~Zhan, ``Structure-aware image fusion,''
  \emph{Optik}, vol. 172, pp. 1--11, 2018.

\bibitem{Liu2018}
Y.~Liu, X.~Chen, Z.~Wang, Z.~J. Wang, R.~K. Ward, and X.~Wang, ``Deep learning
  for pixel-level image fusion: Recent advances and future prospects,''
  \emph{Inform.\ Fusion}, vol.~42, pp. 158--173, 2018.

\bibitem{SIFT}
D.~G. {Lowe}, ``Object recognition from local scale-invariant features,'' in
  \emph{IEEE ICCV}, vol.~2, 1999, pp. 1150--1157.

\bibitem{Ma2019infrared}
J.~Ma, Y.~Ma, and C.~Li, ``Infrared and visible image fusion methods and
  applications: A survey,'' \emph{Inform. Fusion}, vol.~45, pp. 153--178, 2019.

\bibitem{Mertens2007exposure}
T.~Mertens, J.~Kautz, and F.~Van~Reeth, ``Exposure fusion,'' in \emph{Proc.
  Pac. Conf. Comput. Graph. Appl.}, 2007, pp. 382--390.

\bibitem{Mishra2015image}
D.~Mishra and B.~Palkar, ``Image fusion techniques: a review,'' \emph{Int.\ J.\
  Comput.\ Appl.}, vol. 130(9), pp. 7--13, 2015.

\bibitem{brisque}
A.~{Mittal}, R.~{Soundararajan}, and A.~C. {Bovik}, ``Making a “completely
  blind” image quality analyzer,'' \emph{IEEE Signal Process.\ Lett.}, vol.
  20(3), pp. 209--212, 2013.

\bibitem{Moreno2012}
R.~Moreno, L.~Pizarro, B.~Burgeth, J.~Weickert, M.~A. Garcia, and D.~Puig,
  ``Adaptation of tensor voting to image structure estimation,'' in \emph{New
  Develop.\ in the Vis.\ and Process.\ of Tensor Fields}, 2012.

\bibitem{Ochs}
P.~Ochs, Y.~Chen, T.~Brox, and T.~Pock, ``i{P}iano: Inertial proximal algorithm
  for nonconvex optimization,'' \emph{SIAM J. Imaging Sci.}, vol. 7(2), pp.
  1388--1419, 2014.

\bibitem{PAJARES20041855}
G.~Pajares and J.~M. de~la Cruz, ``A wavelet-based image fusion tutorial,''
  \emph{Pat.\ Recogn.}, vol. 37(9), pp. 1855--1872, 2004.

\bibitem{Parisotto_2019}
S.~Parisotto, L.~Calatroni, M.~Caliari, C.-B. Schönlieb, and J.~Weickert,
  ``Anisotropic osmosis filtering for shadow removal in images,'' \emph{Inverse
  Problems}, vol. 35(5), p. 054001, 2019.

\bibitem{Parisotto2018}
S.~Parisotto, L.~Calatroni, and C.~Daffara, ``Digital cultural heritage imaging
  via osmosis filtering,'' in \emph{ICISP 2018}, 2018, pp. 407--415.

\bibitem{Perez2003}
P.~P{\'e}rez, M.~Gangnet, and A.~Blake, ``Poisson image editing,'' \emph{ACM
  Trans.\ Graph.}, vol. 22(3), pp. 313--318, 2003.

\bibitem{RouRou2000}
R.~Rouse and M.~Rouse, \emph{\BIBforeignlanguage{English}{Manuscripts and their
  makers: commercial book producers in medieval Paris, 1200-1500}}.\hskip 1em
  plus 0.5em minus 0.4em\relax Harvey Miller, 2000.

\bibitem{Thies2015real}
J.~Thies, M.~Zollh{\"o}fer, M.~Nie{\ss}ner, L.~Valgaerts, M.~Stamminger, and
  C.~Theobalt, ``Real-time expression transfer for facial reenactment.''
  \emph{ACM Trans.\ Graph.}, vol. 34(6), pp. 183--1, 2015.

\bibitem{Tian2018}
Q.-C. Tian and L.~D. Cohen, ``A variational-based fusion model for non-uniform
  illumination image enhancement via contrast optimization and color
  correction,'' \emph{Signal Processing}, vol. 153, pp. 210 -- 220, 2018.

\bibitem{Vogel2013}
O.~Vogel, K.~Hagenburg, J.~Weickert, and S.~Setzer, ``A fully discrete theory
  for linear osmosis filtering,'' in \emph{SSVM}, 2013, pp. 368--379.

\bibitem{Weickert2013}
J.~Weickert, K.~Hagenburg, M.~Breu{\ss}, and O.~Vogel, ``Linear osmosis models
  for visual computing,'' in \emph{EMMCVPR}, 2013, pp. 26--39.

\bibitem{Zhang2017}
K.~{Zhang}, M.~{Wang}, and S.~{Yang}, ``Multispectral and hyperspectral image
  fusion based on group spectral embedding and low-rank factorization,''
  \emph{IEEE Geosci.\ Remote Sens.\ Lett.}, vol.~55, pp. 1363--1371, 2017.

\bibitem{Zhang2020}
M.~Zhang, S.~Li, F.~Yu, and X.~Tian, ``Image fusion employing adaptive
  spectral-spatial gradient sparse regularization in {UAV} remote sensing,''
  \emph{Signal Processing}, vol. 170, p. 107434, 2020.

\bibitem{Zhao2018}
W.~{Zhao}, H.~{Lu}, and D.~{Wang}, ``Multisensor image fusion and enhancement
  in spectral total variation domain,'' \emph{IEEE Trans.\ Multimedia}, vol.
  20(4), pp. 866--879, 2018.

\end{thebibliography}

\vspace{-3.5em}
\begin{IEEEbiography}
[{\includegraphics[width=1in,height=1.2in,trim=0.25cm 2cm 0.25cm 0cm,clip=true]{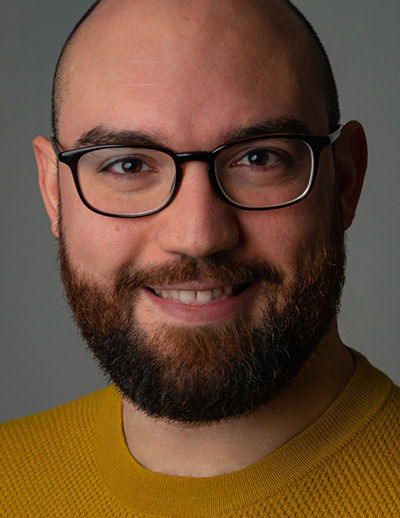}}]{Simone Parisotto}
received the Ph.D.\ degree in Applied Mathematics at the University of Cambridge (UK) in 2019. Before that, he received the B.Sc.\ and the M.Sc.\ in Mathematics at the University of Verona (Italy). He is currently a Post-Doc at the University of Cambridge, in the Cambridge Image Analysis group and the Fitzwilliam Museum, working on mathematical applications for Cultural Heritage.
His research interests focus also on variational and numerical methods for inverse imaging problems.
\end{IEEEbiography}
\vspace{-4.5em}
\begin{IEEEbiography}
[{\includegraphics[width=1in,height=1.2in,trim=1.5cm 7cm 2.5cm 0cm,clip=true]{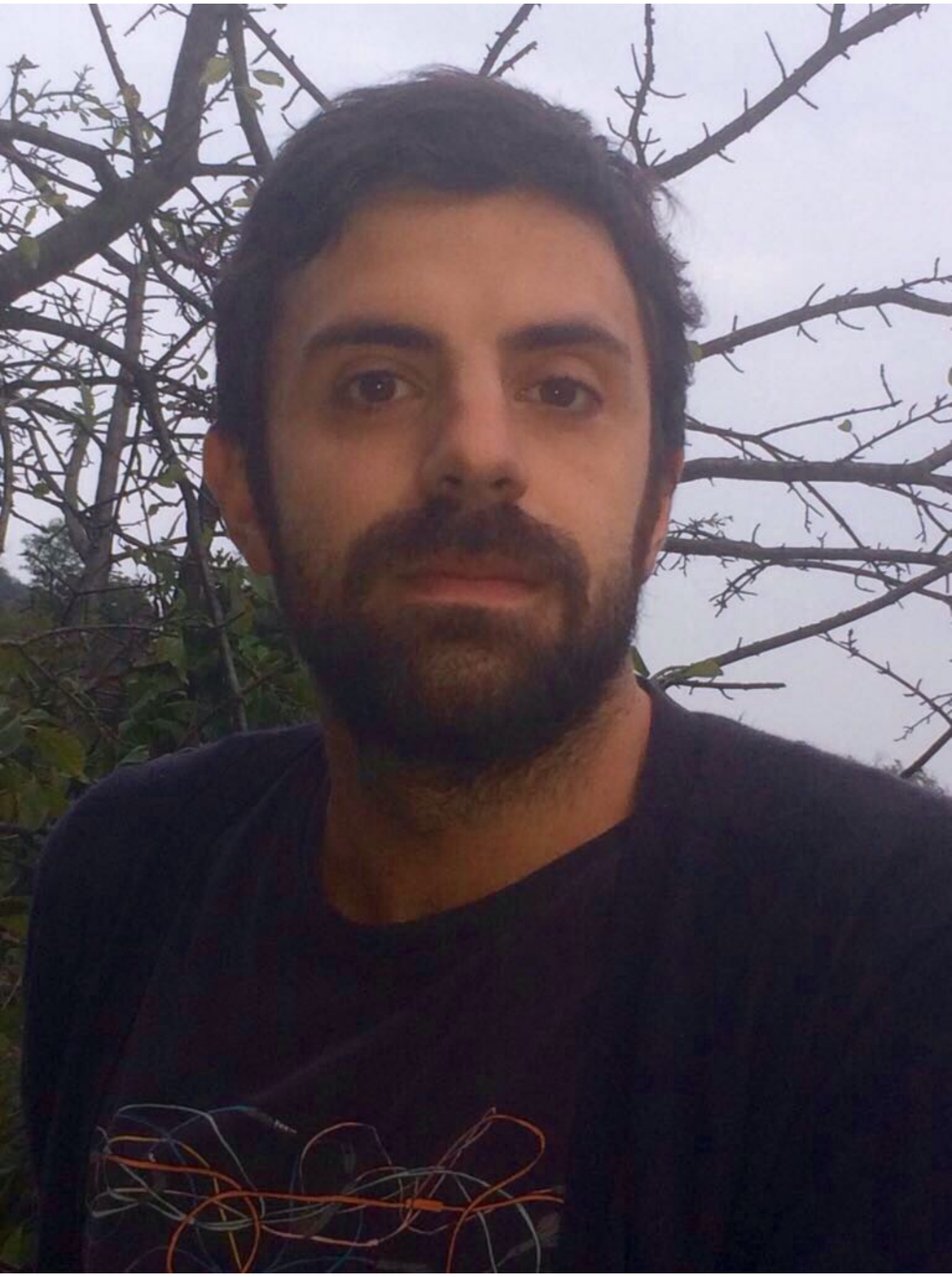}}]{Luca Calatroni}
completed his Ph.D.\ in Applied Mathematics in 2015 as part of the Cambridge Image Analysis research group (UK). After that, he was Post-Doc at the University of Genova (Italy) and \emph{Lecteur Hadamard} fellow at the \'Ecole Polytechnique (France), funded by the FMJH. From October 2019, he is permanent CNRS researcher at the I3S laboratory in Sophia Antipolis, France. His research focuses on variational methods and non-smooth optimization algorithms for imaging.
\end{IEEEbiography}
\vspace{-4.5em}
\begin{IEEEbiography}[{\includegraphics[width=1in,height=1.2in,clip=true]{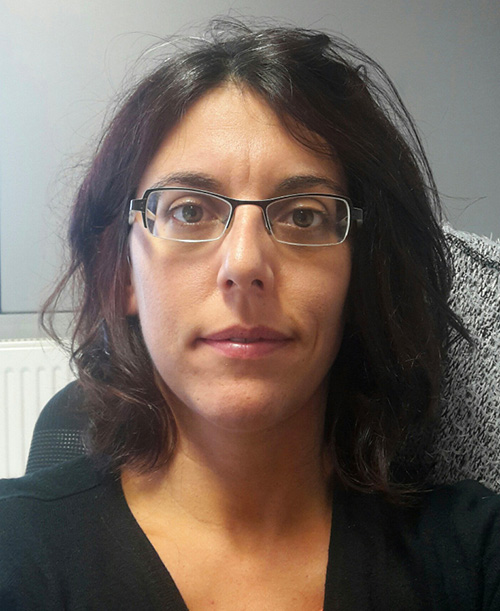}}]{Aurélie Bugeau} received the Ph.D.\ degree in signal processing from the University of Rennes, France, in 2007. Since 2010, she is an Associate Professor in computer science at the University of Bordeaux and at the Laboratoire Bordelais de Recherche en Informatique, where she conducts her research. Her main research interests include patch-based and deep-learning-based methods for image, video and point cloud processing and analysis.
\end{IEEEbiography}
\vspace{-4.5em}
`\begin{IEEEbiography}[{\includegraphics[width=1in,height=1.2in,trim=0.25cm 2cm 0.25cm 0cm,clip=true]{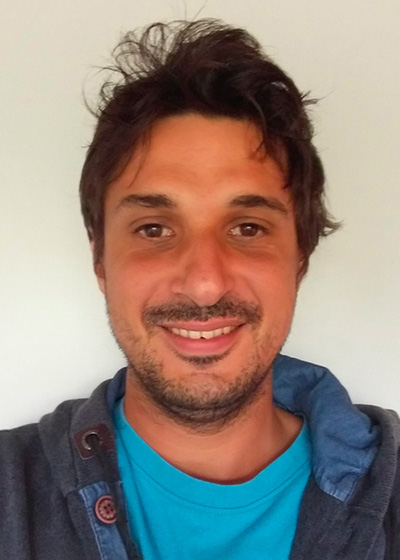}}]{Nicolas Papadakis}
received a M. Eng.\  degree in
applied mathematics from the National Institute of Applied Sciences, Rouen, France (2004), a M.\ Sc in Analysis from
University of Rouen (2004) and  the 
Ph.D.\ degree in applied mathematics from the University of Rennes, France (2007). 
He is currently a full time CNRS Researcher at Institut de Mathématiques
de Bordeaux, France. His research focuses on variational methods and optimal
transportation for image processing applications. 
\end{IEEEbiography}
\vspace{-4.5em}
\begin{IEEEbiography}[{\includegraphics[width=1in,height=1.2in,trim=0cm 2cm 0cm 0cm,clip=true]{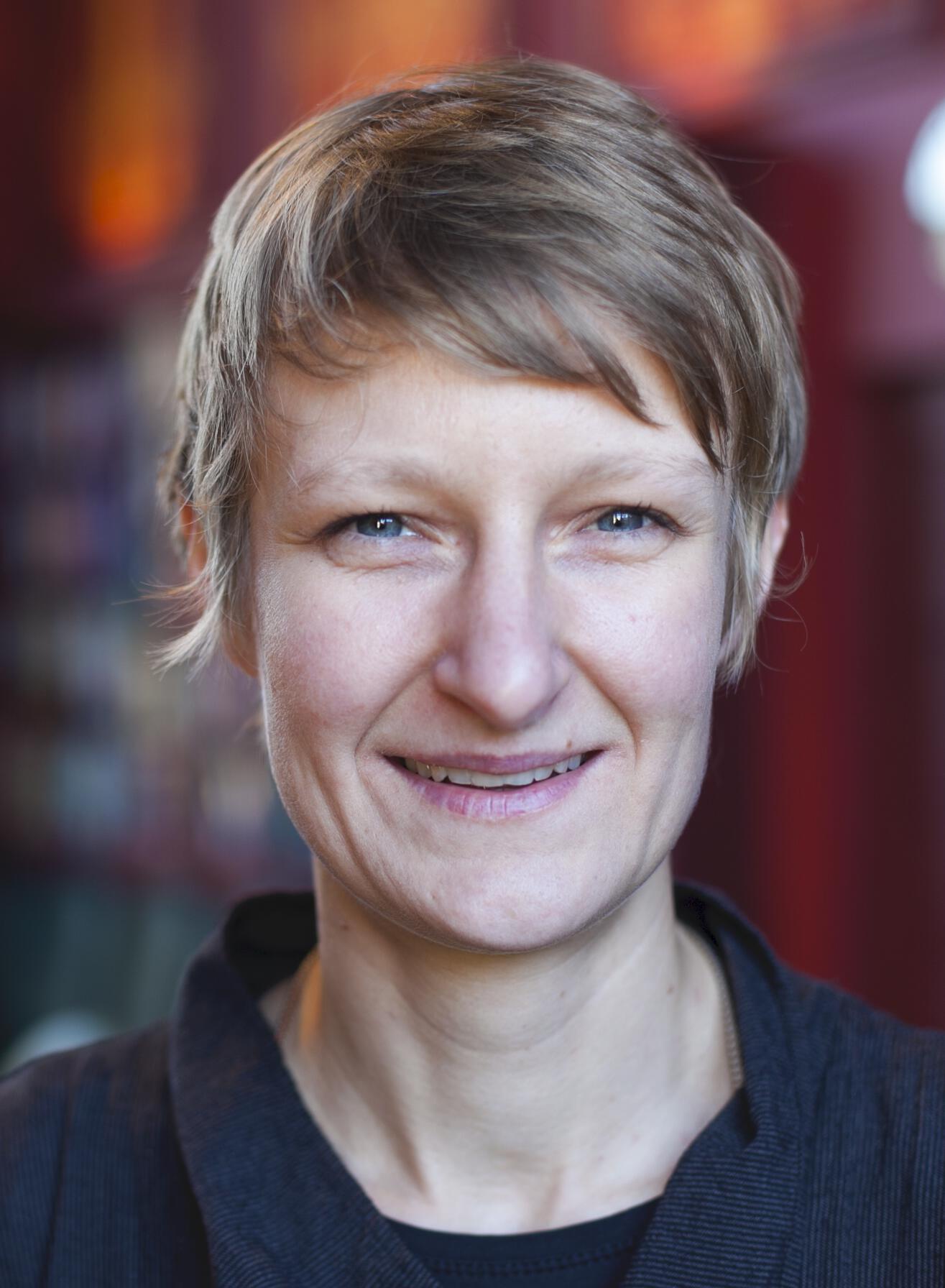}}]{Carola-Bibiane Sch\"onlieb}
received her PhD degree in Mathematics from the University of Cambridge in 2009. After one year of postdoctoral activity at the University of G\"{o}ttingen (Germany), she became a Lecturer at the University of Cambridge in 2010, promoted to Reader in 2015 and promoted to Professor in 2018. Her research interests focus on variational methods, partial differential equations and machine learning for image analysis, image processing and inverse imaging problems.
\end{IEEEbiography}

\end{document}